\documentclass[journal]{IEEEtran}

\usepackage{times}
\usepackage{mathrsfs}
\usepackage{epsfig}
\usepackage{amsmath}
\usepackage{amssymb}
\usepackage{graphicx}
\usepackage{subfigure}
\usepackage{cite}
\usepackage{enumerate}
\usepackage{cases}
\usepackage{multirow}
\usepackage{verbatim}
\usepackage{amssymb}
\usepackage{CJK}
\usepackage{algorithm}
\usepackage{algorithmicx}
\usepackage{algpseudocode}
\usepackage{color}
\usepackage{bm}
\usepackage{booktabs}
\usepackage{hyperref}
\usepackage{amssymb}
\usepackage{array}

\newcommand{\etal}{\textit{et al}.}
\newcommand{\ie}{\textit{i}.\textit{e}.}
\newcommand{\eg}{\textit{e}.\textit{g}.}
\newcommand{\etc}{\textit{etc}}

\begin{document}

\title{CIR-Net: Cross-modality Interaction and Refinement for RGB-D Salient Object Detection}

\author
{
Runmin Cong,~\IEEEmembership{Member,~IEEE,} Qinwei Lin, Chen Zhang, Chongyi Li, Xiaochun Cao,~\IEEEmembership{Senior Member,~IEEE,}\\ Qingming Huang,~\IEEEmembership{Fellow,~IEEE,} and Yao Zhao,~\IEEEmembership{Senior Member,~IEEE}
\thanks{Runmin Cong, Qinwei Lin, Chen Zhang, and Yao Zhao are with the Institute of Information Science, Beijing Jiaotong University, Beijing 100044, China, also with the Beijing Key Laboratory of Advanced Information Science and Network Technology, Beijing 100044, China (e-mail: rmcong@bjtu.edu.cn, lqw22@mails.tsinghua.edu.cn, chen.zhang@bjtu.edu.cn, yzhao@bjtu.edu.cn).}
\thanks{Chongyi Li is with the School of Computer Science and Engineering, Nanyang Technological University, Singapore  (e-mail: lichongyi25@gmail.com).}
\thanks{Xiaochun Cao is with School of Cyber Science and Technology, Shenzhen Campus, Sun Yat-sen University, 518107, China (e-mail: caoxiaochun@mail.sysu.edu.cn).}
\thanks{Qingming Huang is with the School of Computer Science and Technology, University of Chinese Academy of Sciences, Beijing 101408, China, also with the Key Laboratory of Intelligent Information Processing, Institute of Computing Technology, Chinese Academy of Sciences, Beijing 100190, China, and also with Peng Cheng Laboratory, Shenzhen 518055, China (email: qmhuang@ucas.ac.cn).}
}

\markboth{}
{Shell \MakeLowercase{\textit{et al.}}: Bare Demo of IEEEtran.cls for IEEE Journals}
\maketitle

\begin{abstract}
Focusing on the issue of how to effectively capture and utilize cross-modality information in RGB-D salient object detection (SOD) task, we present a convolutional neural network (CNN) model, named CIR-Net, based on the novel cross-modality interaction and refinement.
For the cross-modality interaction, 1) a progressive attention guided integration unit is proposed to sufficiently integrate RGB-D feature representations in the encoder stage, and 2) a convergence aggregation structure is proposed, which flows the RGB and depth decoding features into the corresponding RGB-D decoding streams via an importance gated fusion unit  in the decoder stage.
For the cross-modality refinement, we insert a refinement middleware structure between the encoder and the decoder, in which the RGB, depth, and RGB-D encoder features are further refined by successively using a self-modality attention refinement unit and a cross-modality weighting refinement unit.
At last, with the gradually refined features, we predict the saliency map in the decoder stage.
Extensive experiments on six popular RGB-D SOD benchmarks demonstrate that our network outperforms the state-of-the-art saliency detectors both qualitatively and quantitatively. 
The code and results can be found from the link of \url{https://rmcong.github.io/proj\_CIRNet.html}.
\end{abstract}


\IEEEpeerreviewmaketitle

\section{Introduction} \label{sec1}

\begin{figure}
\centering
	\includegraphics[width=1\linewidth]{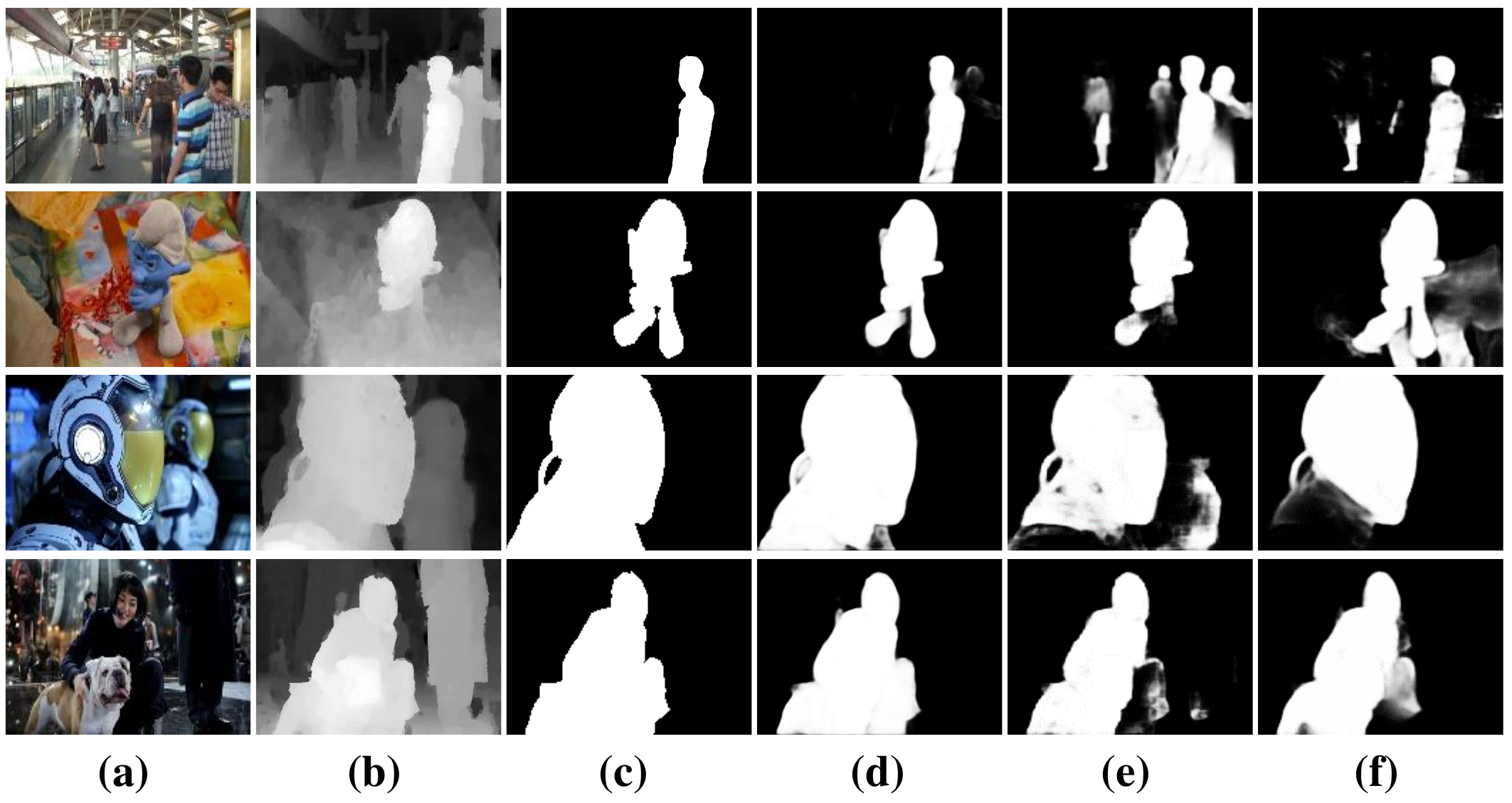}
\caption{Visual examples of different methods. (a) RGB images. (b) Depth maps. (c) Ground truths. (d) Our results. (e)-(f) Saliency maps produced by FRDT \cite{zhang2020feature} and GCPANet \cite{GCPANet}, respectively.}
\label{fig1}
\end{figure}

\IEEEPARstart{W}{HEN} viewing an image, humans are involuntarily attracted by some objects or regions in the image (\eg, the Smurfs in the second image of Fig. \ref{fig1}), which is mainly caused by the human visual attention mechanism, and these objects are called salient objects \cite{cmm_review,RERVIEW,wwg_review}. Simulating this scheme, in the field of computer vision, salient object detection (SOD) is the task of automatically locating the most visually attractive objects or regions in a scene, which has been successfully applied to numerous tasks, such as segmentation \cite{R1,li2020personal,crm/ACMMM20/DMVOS,crm/jbhi22/polyp,crm/tim22/covid,crm/tce22/covid}, retrieval \cite{R3}, enhancement \cite{crm/JEI16/underwater,crm/tip21/underwaterMedium,crm/spl21/underwater,crm/cvpr20/low-light,crm/tmm20/dehazing}, and quality assessment \cite{R2,crm/SPIC21/underwaterIQA}. 

With the development of SOD task research, there are many subtasks, such as co-salient object detection (CoSOD) \cite{crm/ICME18/CoSOD,crm/nips20/CoADNet,crm/tcyb22/glnet}, remote sensing SOD \cite{crm/tgrs19/rsi,crm/tgrs22/RRNet,crm/nc20/rsi,crm/tcyb22/rsi,crm/tip21/DAFNet}, video SOD \cite{crm/tip19/VSOD,li2018flow,ji2021full}, light field SOD \cite{crm/acmmm21/light-field}, have also been developed. In fact, the natural binocular structure of humans can also perceive the depth of field of the scene, and then generate stereo perception. Expressing this depth relationship in the form of an image is a depth/disparity map. In recent years, the development and popularization of depth sensors, especially the rise of affordable and portable consumer depth cameras, has further promoted the applications of RGB-D data, such as depth map super-resolution \cite{gcl2019tip,crm/acmmm21/bridgenet,crm/CVPR21/depthSR}, depth estimation \cite{wang2020eccv}, superpixel segmentation \cite{li2021ins}, and saliency detection \cite{DMRA,crmtc19,A2dele,crmtip18,crmtmm19,crm/tmm22/3DSaliency,crm/tip21/DynamicRGBDSOD}. For the RGB-D images, the RGB image contains abundant details and appearance information (\eg, color, texture, structure, \etc) while the depth map provides some valuable supplementary information (\eg, shape, surface normals, internal consistency, \etc). Recently, more and more studies focus on the introduction of depth cue for the SOD task to effectively suppress the background interference in complex scenes and further completely highlight foreground salient regions. For example, in Fig. \ref{fig1}, the first two images have complex and mussy backgrounds, and the color contrast between the salient object and the background in the fourth image is low. Thus, for the RGB SOD method (\ie, the GCPANet \cite{GCPANet}) shown in the last row of Fig. \ref{fig1}, it is difficult to accurately locate the salient regions with a clean background and a complete structure. In comparison, the RGB-D SOD methods (\eg, the fourth and fifth rows of Fig. \ref{fig1}) can alleviate this problem with the introduction of depth information. Notably, our method has better object positioning ability, completeness preserving ability, and background suppression ability.

The effectiveness of depth map for SOD task has been validated in previous work \cite{PCFN,CPFP,ASIF-Net,cmMS}; however, how to effectively utilize and integrate the RGB information and depth cue is still an open issue. This is because RGB image and depth map belong to different modalities that have different attributes. 
To achieve this, we design the three-stream structure network to fully capture and utilize cross-modality information. Considering the strengths and complementarities of different modalities, through the three-stream structure with independent RGB and depth streams, we can sufficiently preserve the rich information and explore the complementary relations of different modalities, which is beneficial to jointly integrate cross-modality information in the encoder and decoder stage with a more comprehensive and in-depth manner than the two-stream structure. It is manifested in the following two aspects:

1) \textbf{\emph{Cross-Modality Interaction.}} In terms of the cross-modality information, the primary problem we face is how to interact them. Specifically, the purpose is to learn the strengths and complementarities of different modalities, then obtain more comprehensive and discriminative feature representations. Different from the existing cross-modality interaction methods that operated only in the encoder stage \cite{CMW,SSF} or decoder stage \cite{cmMS,S2MA,DMRA}, we dedicate to integrating cross-modality information into both encoder and decoder stages jointly in a more comprehensive and in-depth manner, which sufficiently explores the complementary relations of different modalities. Concretely, in the feature encoder stage, we design a progressive attention guided integration (PAI) unit to fuse cross-modality and cross-level features, thereby attaining  the RGB-D encoder representations. 
In the feature decoder stage, we design an aggregation structure to allow RGB and depth decoder features to flow into the RGB-D mainstream branch and generate more comprehensive saliency-related features. In this structure, the decoder features of the previous layer, the RGB and depth decoder features of the corresponding layer are integrated into confluence decoder features through an important gate fusion (IGF) unit in a dynamic weighting manner. The gradually refined decoder features of the last layer are then used to predict the final saliency map.

2) \textbf{\emph{Cross-Modality Refinement.}} In addition to cross-modality interaction, refining the most valuable information from different modalities is also crucial for RGB-D SOD task. To this end, we insert a refinement middleware between the encoder and the decoder, including the self-modality refinement and cross-modality refinement. 
For the self-modality refinement, in order to reduce the feature redundancy of the channel dimension and emphasize the important location of the spatial dimension, we propose a simple but effective self-modality attention refinement (smAR) unit, which replaces the commonly used progressive interaction \cite{BBS-Net} or feature fusion \cite{cmMS} method with our proposed channel-spatial attention generation. We directly integrate spatial attention and channel attention in the feature map space to generate a 3D attention tensor that is used to refine the single modality features, which not only reduces the computational cost, but also better highlights the important features.
Further, we design a cross-modality weighting refinement (cmWR) unit to refine the multi-modality features by considering cross-modality complementary information and cross-modality global contextual dependencies. Inspired by the non-local model \cite{non-local}, the RGB features, depth features, and RGB-D features are integrated to capture the long-range dependencies among different modalities. Then, we use the integrated features to weight and refine different modality features, thereby obtaining the refined features embedded with cross-modality global context cue, which is important for the perception of global information.

In summary, our method is unique in that the cross-modality interaction and refinement are closely coupled in a comprehensive and in-depth manner. In terms of the cross-modality interaction, for learning the strengths and complementarities of different modalities, we propose the PAI unit in the encoder stage and the IGF unit in the decoder stage to jointly explore the complementary relations of different modalities. In terms of cross-modality refinement, considering the information redundancy of the encoder features and the significance of global context cues for the SOD, we design the pluggable refinement middleware structure to refine the encoder features from the self-modality and cross-modality perspectives.
The main contributions are summarized as follows:
\begin{itemize}
\item We propose an end-to-end cross-modality interaction and refinement network (CIR-Net) for RGB-D SOD by fully capturing and utilizing the cross-modality information in an interaction and refinement manner.
\item The progressive attention guided integration unit and the importance gated fusion unit are proposed to achieve comprehensive cross-modality interaction in the encoder and decoder stages respectively.
\item The refinement middleware structure including the self-modality attention refinement unit and cross-modality weighting refinement unit is designed to refine the multi-modality encoder features by encoding the self-modality 3D attention tensor and the cross-modality contextual dependencies.
\item Without any pre-processing (\eg, HHA \cite{gupta2014learning}) or post-processing (\eg, CRF \cite{krahenbuhl2011efficient}) techniques, our network achieves competitive performance against the state-of-the-art methods on six RGB-D SOD datasets.
\end{itemize}

The rest of this paper is organized as follows. In Section \ref{sec2}, we briefly review the related works of RGB-D SOD. In Section \ref{sec3}, we introduce the technical details of the proposed CIR-Net. Then, the experiments including the comparisons with state-of-the-art methods and ablation studies are conducted in Section \ref{sec4}. Finally, the conclusion is drawn in Section \ref{sec5}.

\section{Related Work} \label{sec2}
\begin{figure}
\centering
	\includegraphics[width=1\linewidth]{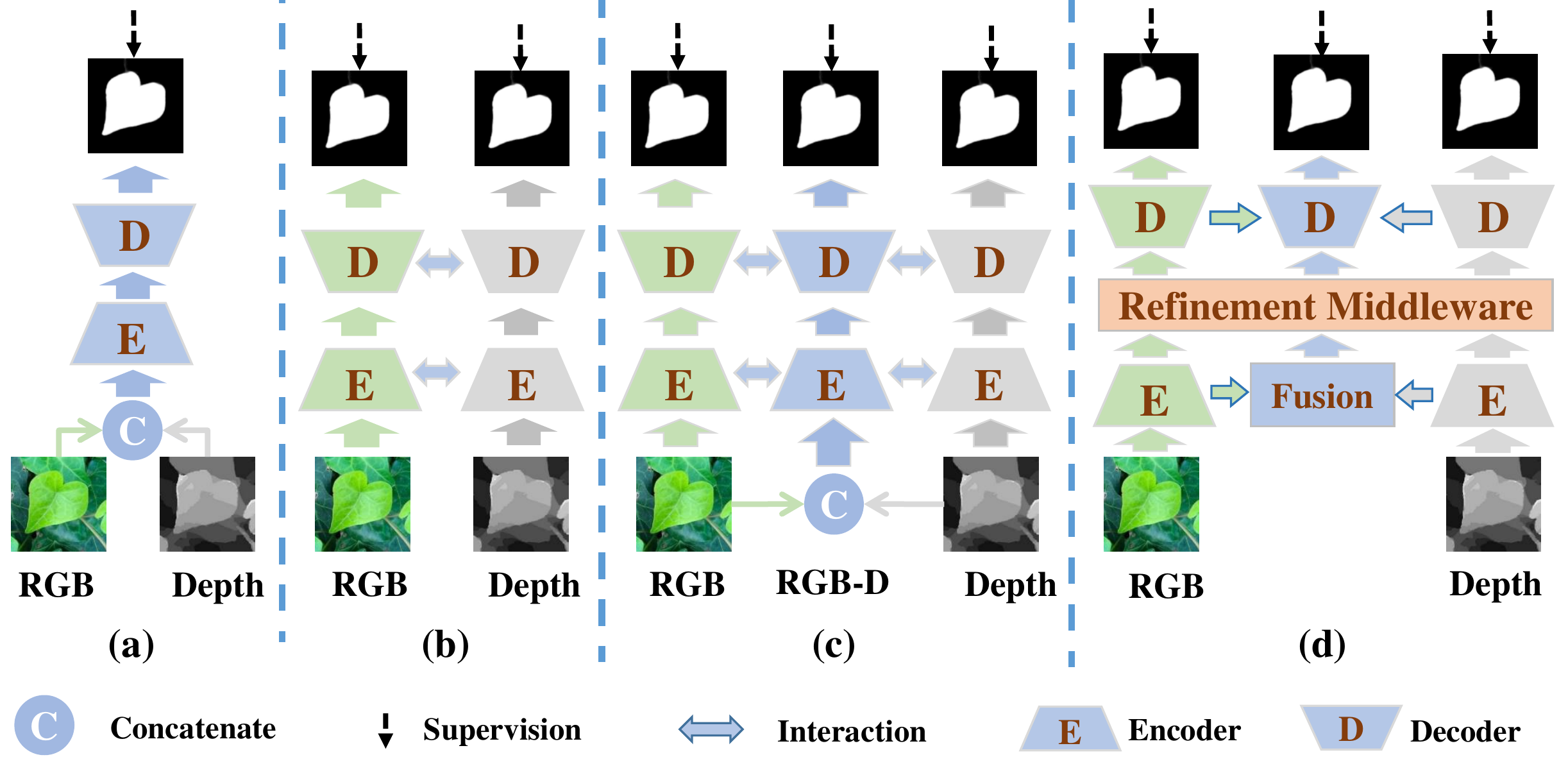}
    \caption{Comparisons among different network structures for RGB-D SOD. (a)-(c) denote the single-stream, two-stream and three-stream structures, respectively. The (d) is the proposed structure in this paper.}
\label{four_mode}
\end{figure}

Different from RGB SOD models \cite{EGNet,zhao2019pyramid,PoolNet,BASNet,crm/tcsvt22/weaklySOD}, depth modality together with RGB appearance are introduced into RGB-D SOD models. In the past ten years, a mass of methods have been proposed, which can be roughly divided into traditional methods \cite{Niu2012,Peng2014,ju2014depth,crm2019tc,DCMC,LBE,ACSD,Song2017,Liang2018} and deep learning-based methods \cite{DF,PCFN,TAN,MMCI,CPFP,cmMS,ASIF-Net,DMRA,A2dele,SSF,S2MA,CMW,DANet,zhang2020feature,fu2021siamese,zhou2019global}. Especially in recent years, the deep learning-based methods have achieved great breakthroughs in the performance of RGB-D SOD.
For the RGB-D SOD task, how to make full use of the cross-modality information and generate more discriminate saliency-related representation is a challenging issue to be addressed \cite{zhou2021rgbd}. In terms of the model structure, the existing works can be roughly divided into single-stream, two-stream and three-stream structures, as shown in Fig. \ref{four_mode}(a)-(c).

For the single-stream models \cite{DANet,UCNet,UInet,CL}, the early feature fusion strategy is commonly adopted, where RGB image and depth map are concatenated into four channels as the input of a network. For examples,  Zhao \etal \cite{DANet} adopted a single-stream encoder to make full use of the representation ability of the pre-trained network, and proposed a real-time and robust salient detection model. Zhang \etal \cite{UCNet,UInet} proposed the first uncertainty-inspired RGB-D SOD model based on conditional variational auto-encoder. Ji \etal \cite{CL} proposed a novel collaborative learning framework that integrated the edge, depth, and saliency collaborators, which is a more lightweight and versatile network due to the free of depth inputs during testing.
However, such models ignore the difference between RGB and depth modalities and lack the comprehensive cross-modality interaction.

The two-stream models \cite{A2dele,cmMS,S2MA,ASIF-Net,DPANet,ICNet, BBS-Net,zhou2020gfnet,zhou2021mrinet} are currently the most widely used structure in RGB-D SOD task, mainly including two independent branches to respectively process RGB and depth modality information and generate cross-modality features in the encoder or decoder stage. 
For example, Li \etal \cite{ASIF-Net} proposed an attention steered interweave fusion network, which progressively and interactively captures cross-modality complementarity via the interweave fusion and weighs the saliency regions by the steering of the deeply supervised attention mechanism. Li \etal \cite{cmMS} adopted the late feature fusion strategy to generate cross-modality representation which combines high-level RGB and depth features of two independent branches in the decoder stage. Zhai \etal \cite{BBS-Net} leveraged the multi-modal and multi-level features to devise a novel cascaded refinement network, and the RGB and depth modalities can be fused in a complementary way. Zhang \etal \cite{CDINet} focused on the roles of RGB and depth modalities in the cross-modality interaction, and presented a discrepant interaction mode, \ie, the RGB modality and the depth modality guide each other interactively. Some studies are taking an interest in the negative impact of low-quality depth maps by controlling, updating, or abandoning the depth information in the two-stream structure \cite{chen2020tip,DPANet,chen2021tip,bai2021circular,HAINet}. 
Chen \etal \cite{DPANet} introduced depth quality perception to control the impact of low-quality depth maps while performing cross-modality interaction in the two-stream structure.
Chen \etal \cite{chen2020tip} estimated an additional high-quality depth map as a complement to the original depth map, and all these depth maps are fed into a selective fusion network to achieve RGB-D SOD.
Chen \etal \cite{chen2021tip} introduced a depth-quality-aware subnet into the two-stream RGB-D SOD structure to locate the most valuable depth regions.

In addition, some studies \cite{TAN,D3Net,zhou2020tsnet} adopted the three-stream network structure for comprehensive cross-modality feature interaction, where RGB, depth, and RGB-D are embedded in three sub-networks for learning and interaction, respectively. For example, Fan \etal \cite{D3Net} designed a gate mechanism to filter out the low-quality depth maps using the decoder results of RGB, depth, and RGB-D branches.

\begin{figure*}[!t]
\centering
\includegraphics[width=1\linewidth]{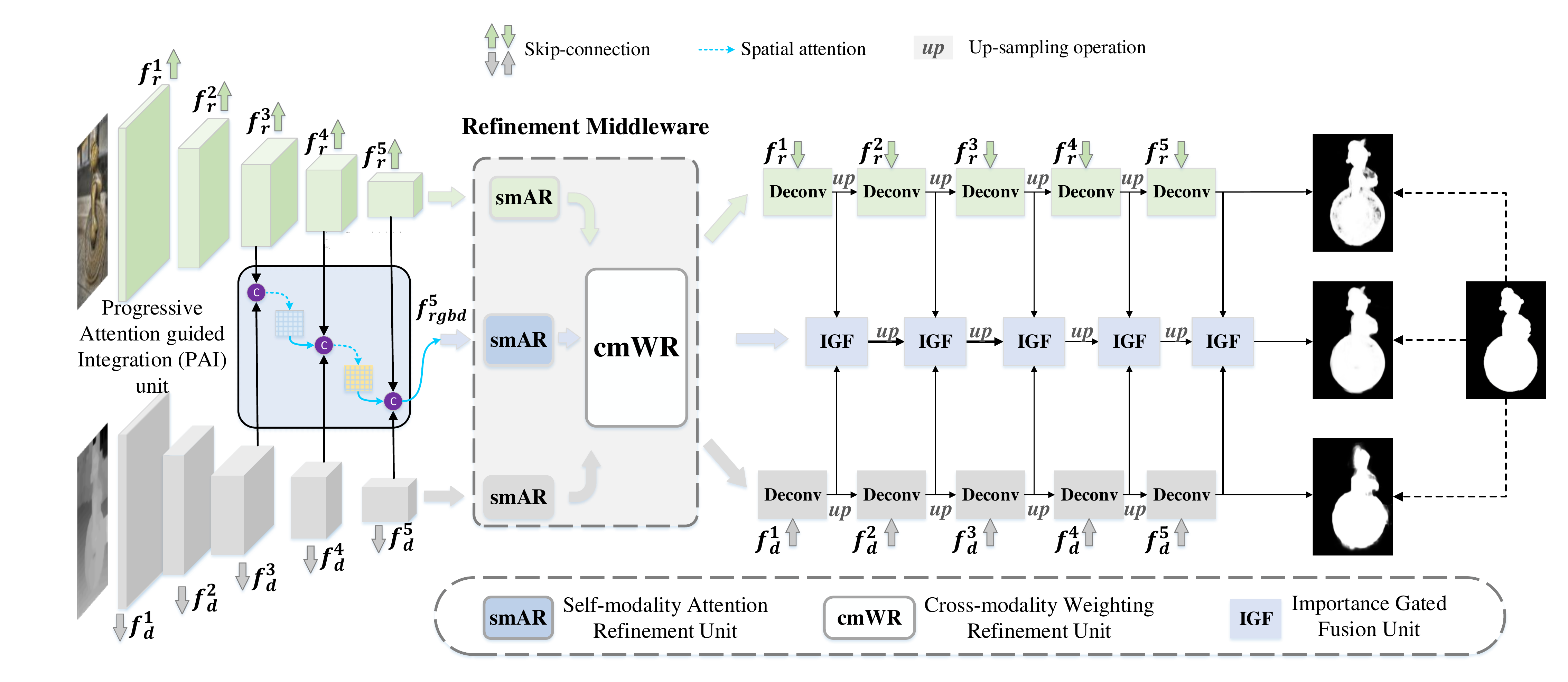}
\caption{The overview of the proposed CIR-Net. The extracted RGB and depth features from the backbone are denoted as $f_r^i$ and $f_d^i$ respectively, where $r$ and $d$ represent RGB and depth streams, and $i\in \{1,2,...,5\}$ indexes the feature level. In the feature encoder, we also use the PAI unit to generate the cross-modality RGB-D encoder features $f_{rgbd}^i~(i\in\{3,4,5\})$. Then, the top-layer RGB, depth, and RGB-D features are embedded into the refinement middleware consisting of a smAR unit and a cmWR unit to progressively refine the multi-modality encoder features in a self- and cross-modality manner. Finally, the decoder features of the RGB branch and depth branch flow into the corresponding RGB-D stream to learn more comprehensive interaction features through an IGF unit in the feature decoder stage. Note that all three branches output a corresponding saliency prediction map, and we use the output of the RGB-D branch as the final result.}
\label{framework}
\end{figure*}


Compared with the existing works, our work differs conceptually from theirs in that: Our proposed network architecture (as shown in Fig. \ref{four_mode}(d)) is a form between two-stream and three-stream networks, and the RGB-D stream is formed by interacting with the high-level features learned by the single-branch network. In this way, the parameters of the network can be reduced, and the RGB-D features can be better established by our designed PAI unit. 
On balance, we classify our network as a three-stream network architecture. This is also the first point that makes our network different from other networks. Second, in addition to the cross-modality feature integration through the PAI unit in the encoder stage, we also perform cross-modality information interaction in the decoder stage to obtain the discriminative saliency prediction features. Considering that the decoder features of RGB and depth streams can further provide effective guidance information (\eg, sharp edge, internal consistency) for RGB-D stream, we design a convergence aggregation structure in the entire decoder stage. 
In this way, we are dedicated to jointly integrating cross-modality information into the encoder and decoder stages in a more comprehensive manner. Third, to better establish the relationship between encoder features and decoder features, we introduce a refinement middleware structure to further highlight the effective information before decoding from the perspective of self-modality and cross-modality. It is worth mentioning that such a middleware structure is pluggable for three-stream networks.

\section{Proposed Method} \label{sec3}

\subsection{Overview}
Fig. \ref{framework} shows the overview of the proposed CIR-Net that is an encoder-decoder three-stream architecture equipped with a refinement middleware between the encoder and the decoder. In what follows, we detail the proposed method.

The feature encoder aims to learn the multi-level three-stream features, \ie, RGB, depth, and RGB-D encoder features.
First, the backbone network (\eg, ResNet50) is used to extract the multi-level features from the input RGB image and depth map, denoted as $f_r^i$ and $f_d^i$, respectively, where $i\in\{1,2,3,4,5\}$ indexes the feature level.
Then, the RGB and depth features at high levels are fed into the proposed progressive attention-guided integration (PAI) unit to generate the cross-modality RGB-D encoder features $f_{rgbd}^i~(i\in\{3,4,5\})$. At this point, the three-stream encoder structure is formed, as shown on the left side of Fig. \ref{framework}.

Considering the information redundancy in the self-modality and the content complementarity in the cross-modality, we introduce a refinement middleware structure to further highlight the effective information before decoding. Specifically, a two-stage refinement mechanism composing of a self-modality attention refinement (smAR) unit and a cross-modality weighing refinement (cmWR) unit is designed to progressively refine the multi-modality top-level encoder features in a self- and cross-modality manner.

In the decoder stage, we devise a novel convergence aggregation structure, in which the corresponding decoder features of the RGB and depth streams flow into the corresponding RGB-D stream to achieve cross-modality interaction. During aggregation, an importance gated fusion (IGF) unit is proposed to integrate the corresponding decoder features of RGB and depth streams and the previous IGF outputs in a dynamic weighting manner. Finally, the output features of the last IGF unit are used to infer the final saliency map.
\begin{figure*}[!t]
	\centering
	\includegraphics[width=1\linewidth]{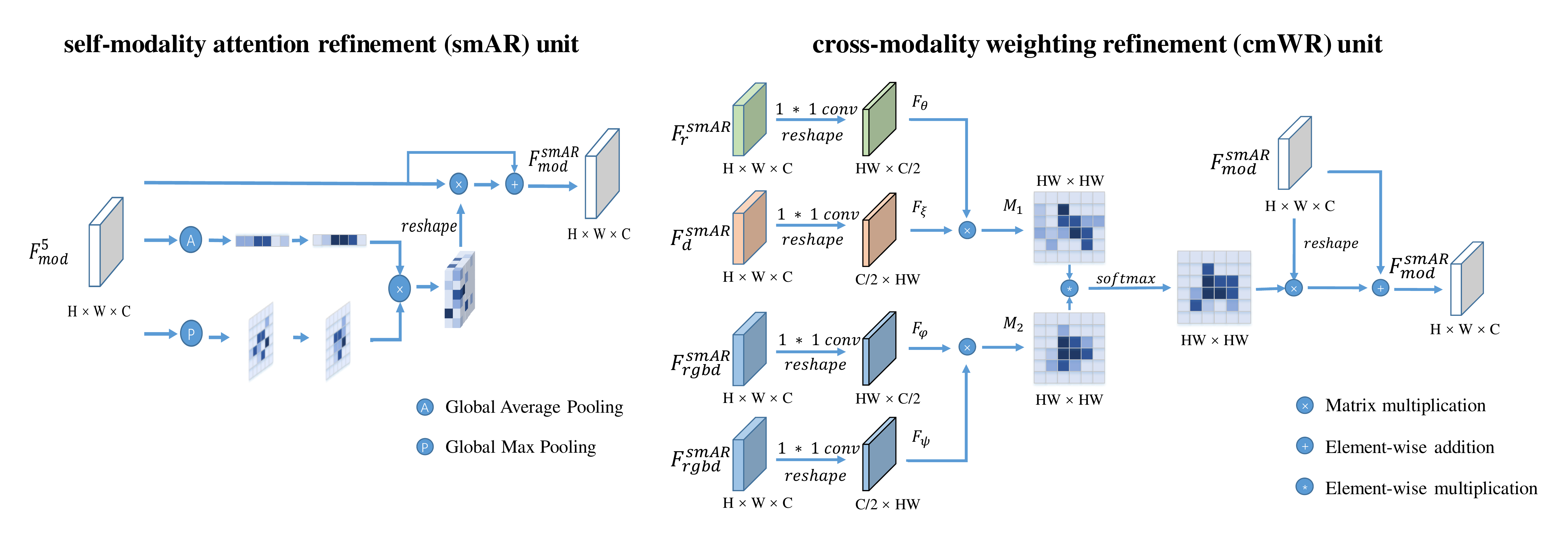}
	\caption{Illustration of the self-modality attention refinement (smAR) unit and the cross-modality weighting refinement (cmWR) unit in the refinement middleware structure.}
	\label{RM}
\end{figure*}
\subsection{Progressive Attention Guided Integration Unit}\label{PAI}
Taking the complementarity and diversity of different modalities into account, effective cross-modality information interaction plays a critical role in the RGB-D SOD task.
For an encoder-decoder network architecture, the existing interaction strategies are mainly designed separately in the encoder stage \cite{CMW,SSF} or decoder stage \cite{cmMS,S2MA,DMRA}. 
In comparison, we design specialized modules in both encoder and decoder stages according to the different interaction purposes. 
To achieve that, two key issues need to be addressed: (1) how to effectively integrate and generate the RGB-D representations based on the multi-level RGB and depth features in the encoder stage, and (2) how the single-modality stream can better collaborate with the RGB-D stream to learn more discriminate saliency-related features and predict more accurate saliency map in the decoder stage. 
To this end, a PAI unit in the encoder stage and an IGF unit in the decoder stage are proposed in our method. The IGF unit will be introduced in Section \ref{IGF}.

Specifically, to effectively integrate the RGB-D representations in the encoder stage, we consider two aspects when designing the PAI unit: (1) sufficient multi-level information fusion, and (2) effective feature selection and highlighting.
For the former, in the encoder stage, considering the fact that the features of different levels contain different information with varying scales, receptive fields, and contents.
Thus, the progressive cross-level fusion strategy is designed to obtain more comprehensive RGB-D representations in a coarse-to-fine manner.
For the latter, although the encoder features contain rich multi-level information, the commonly used fusion strategy (\eg, concat-conv) may introduce information redundancy and easily confuse the feature representations. 
Therefore, for feature selection and enhancement, we introduce the spatial attention scheme to guide the cross-level and cross-modality feature fusion by highlighting the complementary information and suppressing irrelevant redundancy.

First, motivated by the fact that the shallower depth features usually contain too much background noise and the high-level features contain clear information of salient objects but lack details, we choose to generate the initial cross-modality features by combining high-level RGB and depth features and start the combination of features and forward propagation from the third layer, which can be described as:
\begin{align}
	&\tilde{f}_{rgbd}^i=conv([f_{r}^i, f_{d}^i]),i\in \{3,4,5\},
\end{align}
where $f_{r}^i$ and $f_{d}^i$ respectively denote the RGB and depth features at $i^{th}$ encoder level, $[\cdot,\cdot]$ is the channel-wise concatenation operation, and $conv$ represents a convolutional layer followed by a batch normalization (BN) layer and a ReLU activation function.

Then, in order to highlight the complementary information and suppress the irrelevant redundancy in the cross-level and cross-modality fusion, we employ the spatial attention map generated by the previous RGB-D level to guide the current-level feature integration in a progressive manner. Thus, the final RGB-D features at $4^{th}$ and $5^{th}$ levels are updated as:
\begin{align}
	&f_{rgbd}^i = \tilde{f}_{rgbd}^i \odot A^{i-1}+\tilde{f}_{rgbd}^i,i\in \{4,5\} \label{eq2}
\end{align}
\begin{equation}
	\begin{aligned}
		\begin{split}
			A^{i-1}= \left \{
			\begin{array}{ll}
				SA(\tilde{f}_{rgbd}^{i-1} \downarrow),  & i=4\\
				SA({f}_{rgbd}^{i-1} \downarrow),  & i=5\\
			\end{array}
			\right.
		\end{split}
	\end{aligned}
\end{equation}
where $\odot$ is the element-wise multiplication, $A^{i-1}$ denotes the attention map of $(i-1)^{th}$ level, $SA$ is the spatial attention operation \cite{chen2017sca}, and $\downarrow$ denotes the down-sampling operation. 
Note that, considering the inaccurate attention that may happen in some challenging cases, we adopt the residual connection in Eq. (\ref{eq2}) to learn the optimal relationship between the learned features and the original features for effective feature learning. In Section \ref{4.3}, we provide the ablation study to demonstrate the effectiveness of this operation.
Our PAI unit can not only integrate the different modality information, but also encode the different levels' features in a progressive attention weighting manner, thereby generating the RGB-D encoder features.

\subsection{Refinement Middleware}
To transfer more effective encoder features into the decoder stage, we insert a refinement middleware structure as a connecting link between the encoder and decoder to refine the encoder features from the perspectives of self-modality and cross-modality. 
For the design of refinement middleware, we consider two aspects: 
1) the encoder features of each modality contain abundant spatial and channel information while indiscriminate information transmission may increase the difficulty of learning effective feature representations. 
Therefore, we design a smAR unit to suppress the background noises and highlight the important cues from a single modality perspective; 
and 2) considering the strong correlation and complementarity between different modalities where the RGB modality contains figure-ground color contrast and the depth modality contains internal consistency, we design a cmWR unit to capture the long-range dependencies of multiple modalities and refine the modality features from a global perspective.

\subsubsection{Self-modality Attention Refinement Unit}\label{smAR}

After the feature encoding, the obtained RGB, depth, and RGB-D encoder features contain abundant spatial and channel information representing the salient objects. However, there will be redundancy in the single-modality information. Moreover, indiscriminate information transmission may increase the difficulty of feature learning, and even contaminate the inference of subsequent decoding process. Therefore, we design a smAR unit in the refinement middleware to suppress the background noises and highlight the important cues from the perspective of single modality in a new spatial-channel 3D attention manner.

The spatial attention (SA) and channel attention (CA) have been widely used in the existing RGB-D SOD tasks \cite{zhao2019pyramid,BBS-Net,cmMS}, which can be summarized into three forms: (a) Separate utilization. In \cite{zhao2019pyramid}, SA and CA are applied to low-level features and high-level features, respectively. (b) Serial utilization. In \cite{BBS-Net}, the CA is first used to generate the CA-enhanced features, and then the SA is subsequently applied to obtain the final enhanced features. (c) Parallel utilization via feature fusion. In \cite{cmMS}, the CA and SA are respectively used to enhance the same input features, and then the obtained enhanced features are fused to generate the final features. 
The CA and SA in separate utilization are used for different level features, which are not necessarily suitable for all vision tasks. 
However, the serial utilization is sensitive to the order of SA-CA combination, while the way of feature fusion in parallel utilization has some information redundancy in the structural design, and can only enhance the features in one dimension (\ie, spatial or channel) at a time, which increases computational complexity.
To address this issue, we integrate SA and CA into a spatial-channel 3D attention tensor for: 
1) enhancing the robustness via parallel utilization and reducing the computational complexity in a 3D attention manner; 
and 2) refining the single modality features in both spatial and channel dimension simultaneously.

As shown in the left side of Fig. \ref{RM}, the output features of the three encoder branches (\ie, $f^5_{r}$, $f^5_{d}$, and $f^5_{rgbd}$) are embedded into the smAR unit.
We first calculate the CA and SA of the input features in a parallel structure respectively, thereby obtaining the corresponding spatial attention map and channel attention map. Then, we directly fuse them on the attention map space via matrix multiplication to generate the 3D attention tensor. This process can be described as:
\begin{equation}
	A_{3D} = SA(f^5_{mod})\otimes CA(f^5_{mod}) ,
\end{equation}
where $f_{mod}^5$ denotes the each modality features of the top encoder layer, $mod\in\{r,d,rgbd\}$, $SA$ and $CA$ represent the spatial attention \cite{chen2017sca} and channel attention \cite{hu2018squeeze} operations, respectively, and $\otimes$ denotes the matrix multiplication. With the 3D attention tensor, we refine each modality features through a residual connection:
\begin{equation}
	f^{smAR}_{mod} = conv(A_{3D} \odot f^5_{mod} + f^5_{mod}),\label{eq5}
\end{equation}
where $\odot$ is the element-wise multiplication. In Section \ref{4.3}, we provide ablation studies with different attention combinations to demonstrate the effectiveness of our design.


\subsubsection{Cross-modality Weighting Refinement Unit}\label{cmWR}
The smAR unit refines the encoder features in each modality, but does not make full use of the strong correlation and complementarity between different modalities. For example, the RGB modality contains figure-ground color contrast and object texture, and the depth modality provides the internal consistency and spatial relations of the salient objects.
Therefore, inspired by the non-local model \cite{non-local,crm/ijcai20/SR}, we design a novel cmWR unit in the second stage of the refinement middleware to further capture long-range dependencies of multiple modalities.

The details of cmWR are shown in the right side of Fig. \ref{RM}.  The output features of smAR unit $f_{r/d/rgbd}^{smAR} \in \mathbb{R}^{C\times H\times W}$ are embedded into the cmWR unit as the input, where $C$, $H$, and $W$ denote the channel, height, and width of feature maps, respectively.
First, we use bottleneck convolutional layers to half the numbers of channels and map different modalities into a unified feature space, which can be formulated as:
\begin{equation}
	\begin{aligned}
		&F_\theta = W_{\theta}f_{r}^{smAR},\\
		&F_\xi = W_{\xi}f_{d}^{smAR},\\
		&F_\varphi = W_{\varphi}f_{rgbd}^{smAR},\\
		&F_\psi = W_{\psi}f_{rgbd}^{smAR},
	\end{aligned}
\end{equation}
where $W_{\theta}$, $W_{\xi}$, $W_{\varphi}$, and $W_{\psi}$ denote the learnable embedding weights through the bottleneck convolutional layers.

Then, similar to the scaled dot-product attention, the correlation between the RGB features and depth features, and the self-correlation of the RGB-D features are calculated in a pixel-wise manner:
\begin{equation}
	\begin{aligned}
		&M_1 = \texttt{softmax}(F_{\theta}^T \otimes F_\xi),\\
		&M_2 = \texttt{softmax}(F_{\varphi}^T \otimes F_\psi),
	\end{aligned}
\end{equation}
where $\otimes$ is the matrix multiplication, and $\texttt{softmax}$ is the softmax activation function.
$M_1\in\mathbb{R}^{HW\times HW}$ highlights the common response between the RGB and depth modalities, and $M_2\in\mathbb{R}^{HW\times HW}$ models the dependencies of RGB-D modality itself.
The fundamental purpose of dividing $M_1$ and $M_2$ separately is that we want the final similarity interaction to be performed in the RGB-D feature space.

Finally, these two correlation information mapped to the RGB-D modality jointly generate cross-modality global dependency weights to refine the original input features:
\begin{align}
	&f_{mod}^{cmWR} = R(f_{mod}^{smAR})\otimes \texttt{softmax}(M_1 \odot M_2) + f_{mod}^{smAR}, \label{eq8}
\end{align}
where $mod\in\{r,d,rgbd\}$, $\odot$ is the element-wise multiplication, and $R$ reshapes the feature from $\mathbb{R}^{C\times H\times W}$ to $\mathbb{R}^{C\times HW}$.
Through the cross-modality global dependency weights generated by $M_1 \odot M_2$, we refine the original input modality features from a global perspective, which can improve the completeness of the detection result, resulting in higher detection accuracy. 
We conduct various experiments to demonstrate the advantages of the cmWR unit in Table \ref{tab_MiddleWare}.

\subsection{Importance Gated Fusion Unit}\label{IGF}

As we emphasized before, the cross-modality information interaction is essential for RGB-D SOD task. Existing methods usually only interact in a separate encoder or decoder stage, but this is insufficient.
In fact, the encoder and decoder play different roles in feature learning, where the encoder focuses more on general feature extraction while the decoder places extra emphasis on the learning of saliency-related features. Thus, in addition to the cross-modality feature integration through the PAI unit in the encoder stage, we also perform cross-modality information interaction in the decoder stage to obtain the discriminative saliency prediction features. Considering that the decoder features of RGB and depth streams can further provide effective guidance information (\eg, sharp edge, internal consistency) for RGB-D stream, we design a convergence aggregation structure in the entire decoder stage. In detail, the single modality features (\ie, RGB and depth decoder features) at the same level will flow to the corresponding RGB-D stream to learn more comprehensive cross-modality decoder features.
For the convergence aggregation structure, we face a challenging problem, \ie, how to effectively select the most valuable information from the afflux streams, because the direct and equal combination of different modality information may be uncontrollable and miscellaneous. To solve this issue, we design an IGF unit to learn an importance map $P^i$, which is used to selectively control the influence of different modalities in a dynamic weighting manner, as shown in Fig. \ref{figIGF}. In this way, the IGF unit can determine the contribution of supplementary information of different modalities during cross-modality information interaction.
Furthermore, with such learnable important weights, our network is somewhat resistant to situations where certain modal features are invalid, such as low-quality depth maps.

\begin{figure}[!t]
	\centering
	\includegraphics[width=1\linewidth]{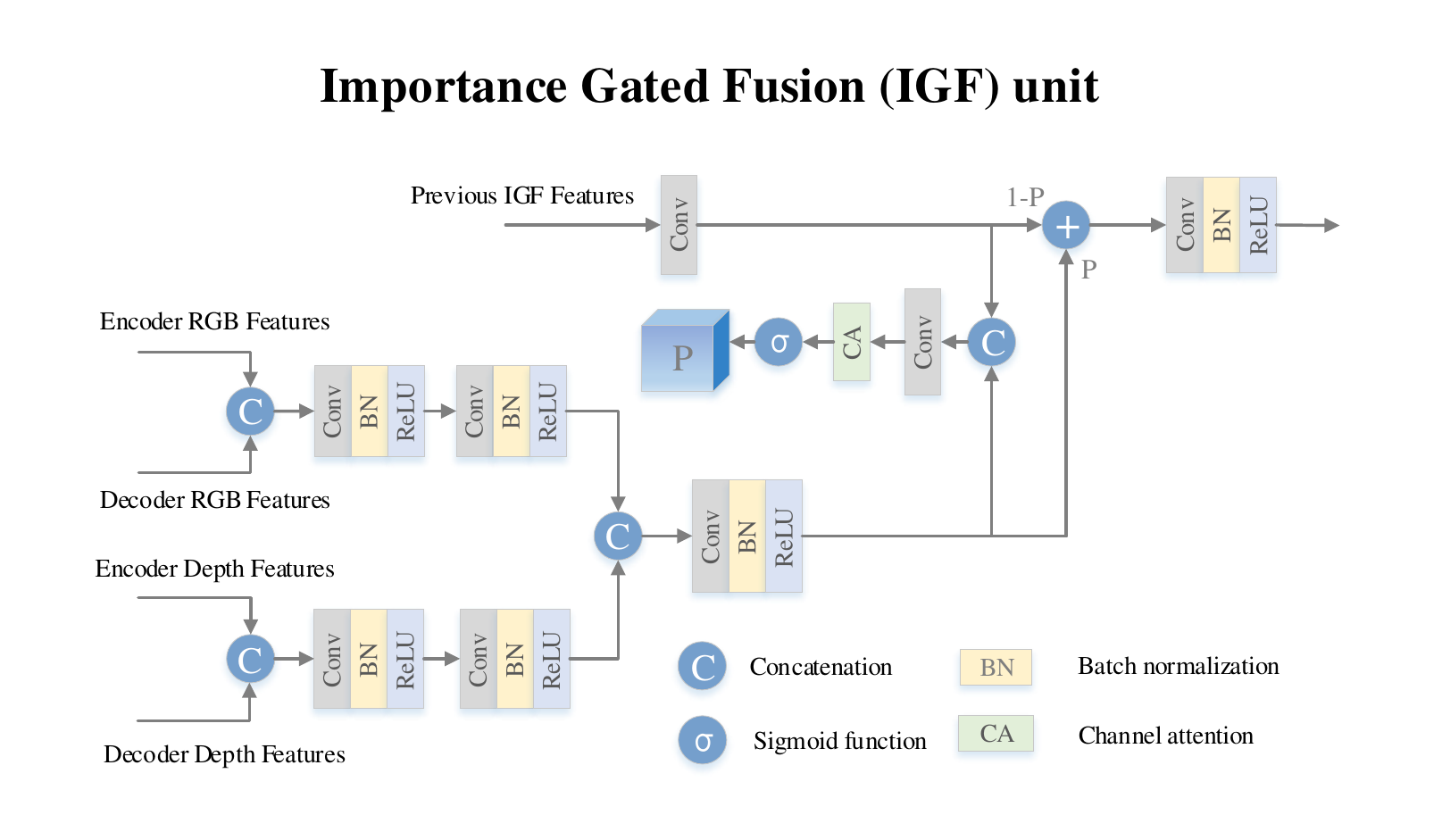}
	\caption{Architecture of the importance gated fusion (IGF) unit.}
	\label{figIGF}
\end{figure}
First, the RGB decoder features and the depth decoder features are fused with the corresponding skip-connection encoder features via two convolutional layers, thus attaining the fused decoder features.
Then, the fused decoder features of the RGB and depth streams are concatenated to obtain the RGB-D decoder features $H^i$.
Finally, the previous IGF features $f_{IGF}^{i+1}$ and the RGB-D decoder features $H^i$ are combined into the current IGF outputs through the learnable importance weight:
\begin{align}
	f_{IGF}^i = conv(P^i\odot H^i+(1-P^i)\odot f_{IGF}^{i+1}\uparrow),
\end{align}
where $f_{IGF}^{i+1}$ denotes the IGF output features at the $(i+1)^{th}$ decoder level, $i\in \{5,4,...,1\}$, $\uparrow$ represents the up-sampling operation, and $P^i$ is the learnable importance weight, which measures the importance of the RGB-D decoder features in the fusion process.
Specifically, we first concatenate two features (\ie, $H^i$ and $f_{IGF}^{i+1}$), and then apply $1\times 1$ convolution to reduce the numbers of channels, producing the features $U^i$. Next, we use the channel attention with the sigmoid activation function to obtain the importance map $P^i\in\mathbb{R}^{C\times H\times W}$:
\begin{align}
	P^i = \sigma(CA(U^i))=\sigma(CA(conv([H^i, f_{IGF}^{i+1}\uparrow]))),
\end{align}
where $CA$ denotes the channel-wise attention \cite{hu2018squeeze}, and $\sigma$ is the sigmoid activation function. The importance map determines the contribution of supplementary information of different modalities at the $i^{th}$ decoder level.

\begin{figure}[!t]
\centering
	\includegraphics[width=0.8\linewidth]{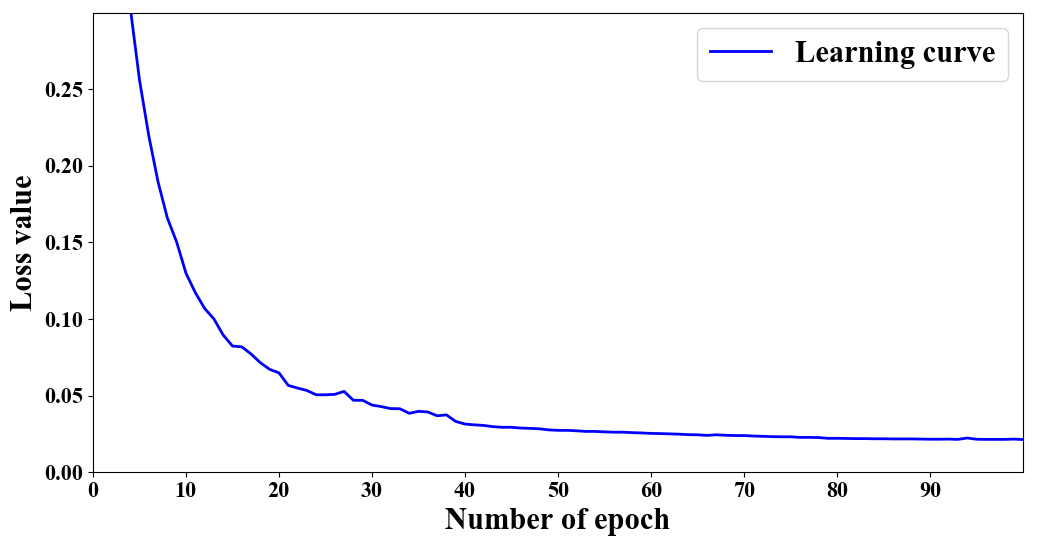}
\caption{Learning curve of our network on the training stage.}
\label{loss_curve}
\end{figure}

\subsection{Loss Function}
In our CIR-Net, the last layer of decoder features of the three streams are used to separately predict the corresponding saliency maps, which are denoted as $S^r$, $S^d$, and $S^{rgbd}$. For network training, we employ the binary cross-entropy (BCE) loss function to optimize the RGB, depth, and RGB-D streams simultaneously. The final loss function is defined as:
\begin{equation}
	Loss = \ell_{bce}(S^r,G)+\ell_{bce}(S^d,G)+\ell_{bce}(S^{rgbd},G),
\end{equation}
where $G$ is the ground truth, and $\ell_{bce}$ is the BCE loss as defined in \cite{crm/tcyb22/glnet,crm/tgrs22/RRNet}.
During the testing phase, we only utilize the prediction of RGB-D stream as the final saliency map.

\section{Experiments}\label{sec4}
We first describe the six RGB-D SOD benchmark datasets and three commonly used evaluation metrics, then introduce the implementation details of the proposed model.
After that, the comparisons with 15 state-of-the-art CNN-based methods are conducted.
Finally, we conduct a series of ablation studies to validate the effectiveness of our proposed modules.

\subsection{Experimental Settings}\label{4.1}

\subsubsection{Benchmark Datasets}
We conduct experiments on six popular RGB-D SOD benchmark datasets, including STEREO797 \cite{Niu2012}, NLPR \cite{Peng2014}, NJUD \cite{ju2014depth}, DUT \cite{DMRA}, LFSD \cite{piao2019saliency} and SIP \cite{D3Net}.
NJUD \cite{ju2014depth} contains 1985 RGB-D images and corresponding manually labeled ground truth.
The images are collected from the Internet and stereo movies with diverse objects and complex scenarios, and the depth maps are estimated from the stereo images.
NLPR \cite{Peng2014} consists of 1000 multiple salient objects RGB-D images, where the depth maps are captured by the Kinect with a resolution of $640 \times 480$.
STEREO797 \cite{Niu2012} includes 797 stereoscopic images collected from the Internet, and the depth maps are estimated from the stereo images.
DUT \cite{DMRA} contains 1200 paired RGB-D images captured by a Lytro camera with a resolution of $600 \times 400$.
LFSD \cite{piao2019saliency} is a small-scale dataset including 100 small-resolution RGB-D images, where the depth maps are captured via a Lytro light field camera.
SIP \cite{D3Net} includes 929 RGB-D images with a high-resolution of $744 \times 992$.

\begin{figure*}[!t]
	\centering
	\includegraphics[width=1\linewidth]{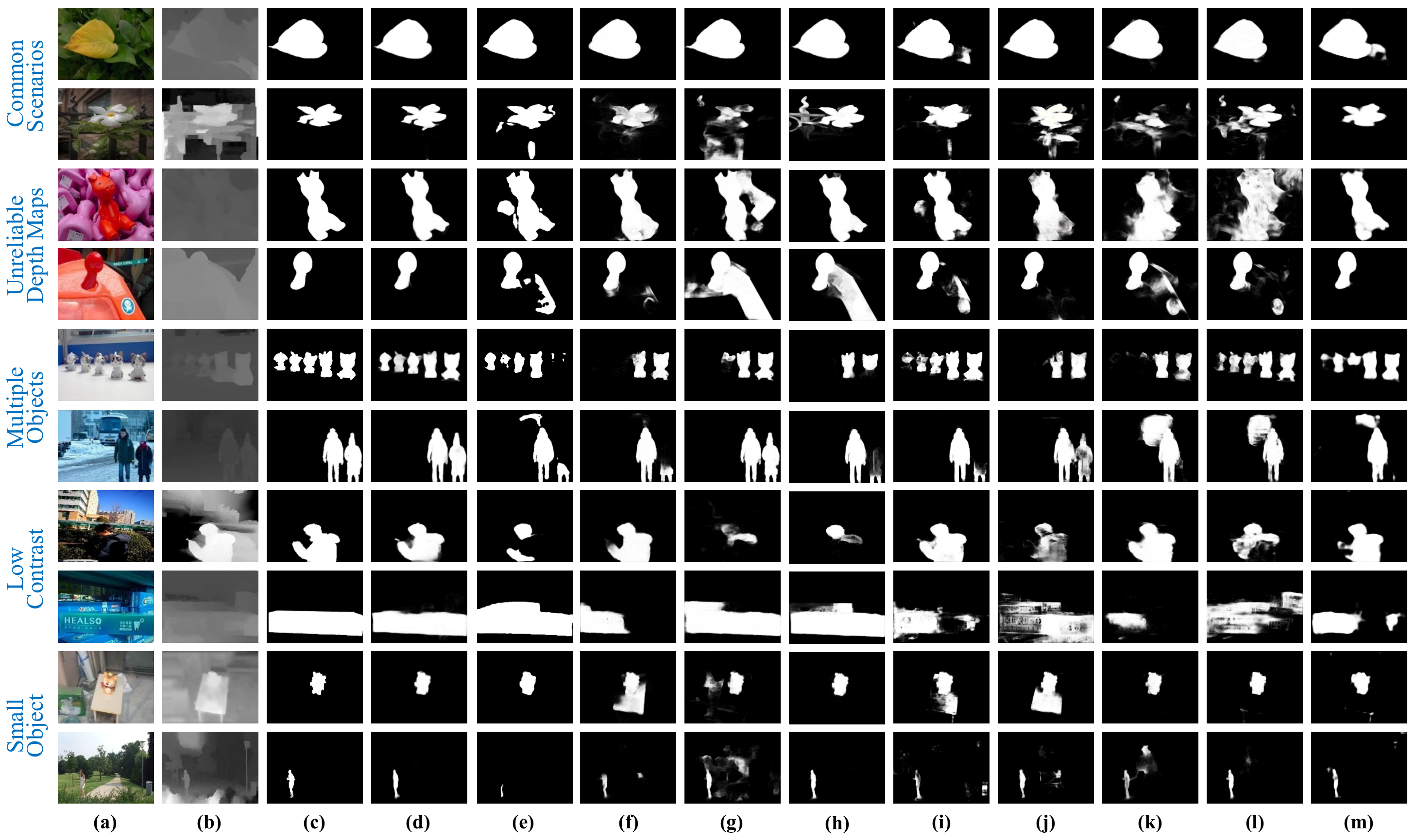}
	\caption{Visual examples of different methods. (a) RGB image. (b) Depth map. (c) GT. (d) Ours. (e) A2dele \cite{A2dele}. (f) DANet \cite{DANet}. (g) S2MA \cite{S2MA}. (h) PGAR \cite{PGAR}. (i) FRDT \cite{zhang2020feature}. (j) JL-DCF \cite{JL-DCF}. (k) D3Net \cite{D3Net}. (l) BiANet \cite{BiANet}. (m) DMRA \cite{DMRA}.  Our method outperforms other SOTA algorithms in various scenes, including common scenarios (line 1 and 2), unreliable or confusing depth maps (line 3 and 4), multiple objects (line 5 and 6), low contrast (line 7 and 8), and small objects (line 9 and 10).}
	\label{visual}
\end{figure*}

\subsubsection{Evaluation Metrics}
To quantitatively evaluate the performance of the proposed method, precision-recall (P-R) curves, F-measure ($F_\beta$) \cite{Niu2012}, Mean Absolute Error (MAE) score \cite{RERVIEW}, and S-measure ($S_m$) \cite{fan2017structure} are employed.
By thresholding the saliency map from 0 to 255, the precision and recall scores can be calculated by comparing the binary mask with the corresponding ground truth, and the variation tendency of different precision and recall scores can be drawn in a precision-recall curve.

\begin{figure*}[!t]
  \centering
    \includegraphics[width=1\linewidth]{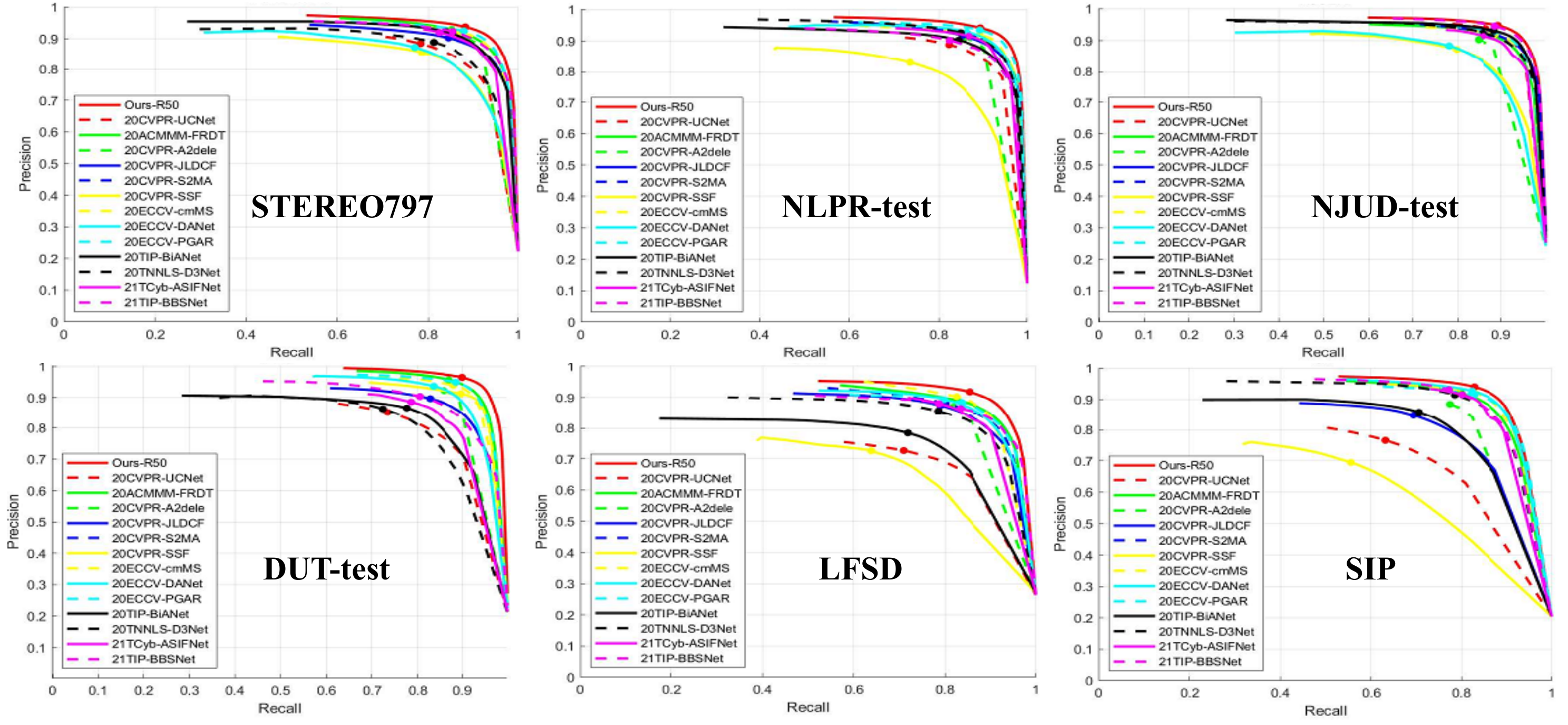}
  \caption{The P-R curves of different methods. Our model (\ie, the red solid line) achieves both higher precision and recall scores against other compared methods over all six benchmark datasets.}
  \label{PR}
  \end{figure*}
\begin{table*}[!h]
\begin{center}
\caption{Quantitative comparison resutls in terms of S-measure, maximum F-measure, and MAE on six  RGB-D benchmark datasets. The bold indicates the best result under each case. The type indicates whether the method is single-stream, two-stream, or three-stream. V16, V19 and R50 denote VGG16, VGG19 and ResNet50, respectively.} \label{tab1}
\renewcommand\arraystretch{1.4}
\setlength{\tabcolsep}{1.0mm}{
\resizebox{18cm}{!}{
\begin{tabular}{c|ccc|ccc|ccc|ccc|ccc|ccc|ccc}
\hline\hline
 &\multirow{2}{*}{year} &\multirow{2}{*}{type}  &\multirow{2}{*}{backbone}   &\multicolumn{3}{c|}{\textbf{STEREO797\cite{Niu2012}}}  &\multicolumn{3}{c|}{\textbf{NLPR-test\cite{Peng2014}}}  & \multicolumn{3}{c|}{\textbf{NJUD-test\cite{ju2014depth}}}  &\multicolumn{3}{c|}{\textbf{DUT-test\cite{DMRA}}}     &\multicolumn{3}{c|}{\textbf{LFSD\cite{piao2019saliency}}}    &\multicolumn{3}{c}{\textbf{SIP\cite{D3Net}}}   \\ \cline{5-22}
& &  &   &$ F_{\beta}\uparrow$ &$ S_{\alpha}\uparrow$  & MAE$ \downarrow$  &$ F_{\beta}\uparrow$ &$ S_{\alpha}\uparrow$  & MAE$ \downarrow$    &$ F_{\beta}\uparrow$ &$ S_{\alpha}\uparrow$  & MAE$ \downarrow$ &$ F_{\beta}\uparrow$ &$ S_{\alpha}\uparrow$  & MAE$ \downarrow$ &$ F_{\beta}\uparrow$ &$ S_{\alpha}\uparrow$  & MAE$ \downarrow$   &$ F_{\beta}\uparrow$ &$ S_{\alpha}\uparrow$  & MAE$ \downarrow$\\ \hline 
DMAR\cite{DMRA}	&2019   &Two    &V19    &.8861 	  &.8858	  &.0474	    &.8749	  & .8892
&.0339  &.8883  &.8804  &.0521  &.8975  &.8879  &.1126  &.8523  &.8393  &.0830  &.8209  &.8060  &.0857\\ 
FRDT\cite{zhang2020feature} &2020	&Two    &V19    &.8987  &.9004  &.0428  &.8976  &.9129  &.0290  &.8982  &.8992  &.0467  &.9263  &.9159  &.0362  &.8555  &.8498  &.0809  &.8714  &.8671  &.0604 \\ 	
SSF\cite{SSF}   &2020	  &Two  &V16    &.8903   &.8920  &.0449  &.8986  &.9141  &.0259  &.9000  &.9002  &.0422  &.9242  &.9157  &.0340  &.8626  &.8495  &.0751  &.8797  &.8737  &.0531\\
S2MA\cite{S2MA} &2020	  &Two  &V16	  &.8158   &.8424   &.0746   &.9017   &.9155   &.0298   &.8888   &.8943   &.0532   &.8997   &.9031   &.0440   &.8310   &.8292   &.1018 &.6317  &.6919   &.1429 \\
A2dele\cite{A2dele} &2020 &Two  &V16    &.8864   &.8868   &.0431   &.8815   &.8979   &.0285   &.8733   &.8704   &.0510   &.8923   &.8864   &.0426   &.8280   &.8258   &.0839   &.8337   &.8287   &.0699 \\
JL-DCF\cite{JL-DCF} &2020  &Single &V16    &.8740   &.8855   &.0509   &.8915   &.9097   &.0295   &.9042   &.9022   &.0413   &.8612   &.8758   &.0556   &.8217   &.8171   &.1031   &.7774   &.7924   &.0967 \\
PGAR\cite{PGAR}     &2020  &Two &V16     &.9008   &.9066   &.0422   &.9153   &.9297   &.0245   &.9068   &.9089   &.0422   &.9171   &.9136   &.0372   &.8390   &.8444   &.0818   &.8759   &.8761   &.0552 \\
DANet\cite{DANet}   &2020  &Single &V16     &.8199   &.8410   &.0712   &.9013   &.9152   &.0283   &.8927   &.8971   &.0463   &.8954   &.8894   &.0465   &.8417   &.8375   &.1031   &.8740   &.8784   &.0537 \\
cmMS\cite{cmMS}     &2020  &Two &V16    &.8971   &.8999     &.0429    &.9031    &.9176    &.0277  &.9034    &.9051   &.0432   &.9090   &.9070   &.0405   &.8623   &.8491   &.0792   &.8814   &.8755   &.0560 \\
BiANet\cite{BiANet} &2020  &Two &V16     &.8844   &.8882   &.0497   &.8764   &.9000   &.0325   &.9121   &.9119   &.0399   &.8156   &.8368   &.0745   &.7287   &.7422   &.1340   &.7869   &.8030   &.0912 \\
D3Net\cite{D3Net}   &2020    &Three   &V16    &.8495   &.8687   &.0578   &.8969   &.9117   &.0296   &.8996   &.9002   &.0465   &.7855   &.8152   &.0848   &.8062   &.8167   &.1023   &.8611   &.8603   &.0631 \\
UCNet\cite{UCNet}   &2020  &Two &V16     &.8355   &.8243   &.0650   &.9034   &.9196   &.0250   &.8860   &.8970   &.0430   &.7785   &.8064   &.0902   &.8595   &.8564   &.0738   &.8792   &.8675   &.0514 \\
ASIFNet\cite{ASIF-Net}  &2021   &Two    &V16     &.8800   &.8820   &.0485   &.8907   &.9079   &.0295   &.8886   &.8902   &.0472   &.8245   &.8396   &.0724   &.8602   &.8520   &.0809   &.8613   &.8594   &.0603 \\
BBSNet\cite{BBS-Net}    &2021   &Two    &R50     &.8778   &.8987   &.0429   &.9153   &.9302   &.0242   &.9194   &.9208   &.0352   &.9237   &.9204   &.0349   &.8689   &.8696   &.0723   &.8945   &.8866   &.0520 \\
UCNet*\cite{UInet}      &2021   &Two    &R50     &-   &-   &-   &.9093   &.9222   &.0234   &.8930   &.9020   &.0390   &.8553   &.8643   &.0561   &.8589   &.8558   &.0713   &.8891   &.8820   &\textbf{.0452}  \\
\hline
\multirow{2}{*}{CIR-Net} &\multirow{2}{*}{--}    &\multirow{2}{*}{Three}  &V16 &.9059   &.9085   &.0413   &.9177   &.9296   &.0227   &.9163   &.9167   &.0382   &.9339   &.9283   &.0320   &.8730   &.8691   &.0682   &.8866   &.8801   &.0553   \\
& & &R50 &\textbf{.9139}   &\textbf{.9166}   &\textbf{.0377}   &\textbf{.9241}   &\textbf{.9334}   &\textbf{.0227}   &\textbf{.9277}   &\textbf{.9250}   &\textbf{.0350}   &\textbf{.9376}   &\textbf{.9324}   &\textbf{.0288}   &\textbf{.8828}   &\textbf{.8753}   &\textbf{.0677}   &\textbf{.8959}   &\textbf{.8884}   &.0523 \\\hline\hline
\end{tabular}}}
\end{center}
\end{table*}
F-measure is a widely used comprehensive evaluation metrics by considering both precision and recall scores, which is defined as:
  \begin{equation}
      F_{\beta} = \frac{(1+\beta)^{2} \cdot Precision \times Recall}{\beta^{2} \times Precision + Recall},
  \end{equation}
where $Precision$ and $Recall$ respectively represent the precision score and recall score, and $\beta^2$ is set to 0.3 for emphasizing the precision as suggested in \cite{Niu2012}.

The MAE score calculates the average pixel-wise absolute difference between the predicted saliency map $S$ and the corresponding ground truth $G$, which is denoted as:
  \begin{equation}
      \text{MAE} = \frac{1}{H \times W}\sum_{y=1}^{H}\sum_{x=1}^{W}\vert S(x,y) - G(x,y) \vert,
  \end{equation}
where $H$ and $W$ represent the height and width of the image, respectively.

S-measure denotes the structural similarity between the predicted saliency map and the corresponding ground truth:
  \begin{equation}
      S_m = \alpha \times S_o + (1 - \alpha) \times S_r,
  \end{equation}
where $\alpha{}$ is set to 0.5 to balance the region similarity $S_r$ and object similarity $S_o$ as suggested in \cite{fan2017structure}.

\subsubsection{Implementation Details}
Following \cite{A2dele,cmMS}, we adopt $1485$ samples from NJUD dataset, $700$ samples from NLPR dataset, and $800$ samples from DUT dataset as the training data.
The remaining samples in these three datasets and the rest three datasets are used as testing datasets.
During training, the random flipping, rotating and multi-scale input are adopted for data augmentation.
During the training phase, the training samples are randomly resized to $128\times 128$, $256\times256$, and $352\times352$. 
In the interference stage, the images are resized to $352\times 352$ and then fed into the network to obtain saliency prediction without any other post-processing or pre-processing techniques.
We report the experimental results using ResNet50 and VGG16 as backbone networks, initialized by the pre-trained parameters on ImageNet \cite{imagenet}.\footnote{Unless otherwise stated, the results in this paper are obtained with ResNet as the backbone network.}
The Adam algorithm is used to optimize our network with a batch size of $16$, and the initial learning rate $1e$-$4$ is divided by $5$ every $40$ epochs.
Our network is implemented in PyTorch and accelerated by two NVIDIA $2080$Ti GPUs. We also implement our network by using the MindSpore Lite tool\footnote{\url{https://www.mindspore.cn/}}. In order to show the training process of our model more clearly, we report the learning curve of our network in Fig. \ref{loss_curve}. It takes around 4 hours to optimize our network.
The inference time of our method is $0.07$ second for an image with the size of $352\times352$. 

\subsection{Comparison with the State-of-the-art Methods}\label{4.2}

We compared the proposed model with $15$ state-of-the-art CNN-based RGB-D SOD methods, including DMRA \cite{DMRA}, FRDT \cite{zhang2020feature}, SSF \cite{SSF}, S2MA \cite{S2MA}, A2dele \cite{A2dele}, JL-DCF \cite{JL-DCF}, PGAR \cite{PGAR}, DANet \cite{DANet},  cmMS \cite{cmMS}, BiANet \cite{BiANet}, D3Net \cite{D3Net}, UCNet \cite{UCNet}, ASIF-Net \cite{ASIF-Net}, BBSNet \cite{BBS-Net},  and UCNet* \cite{UInet} (the extension version of UCNet).
For fair comparisons, all the saliency maps are generated by the released code under the default settings or are provided by the authors directly.

To further illustrate the outperformance of our proposed method, we provide some visualization comparison results of different methods in Fig. \ref{visual}.
From it, we can clearly see that our proposed model achieves superior performance, which achieves accurate location and complete structure of the salient objects.
For quantitative evaluations, we report the P-R curves of different methods on six benchmark datasets, which is shown in Fig. \ref{PR}. The closer the P-R curve is to $(1, 1)$, the better the algorithm performance.
As visible, our model (\ie, the red solid line) achieves both higher precision and recall scores against other compared methods over all six benchmark datasets.
Moreover, as shown in Table \ref{tab1}, our method achieves the best performance except for the MAE metric on the SIP dataset, which also demonstrates the effectiveness and superiority of the proposed method.
For example, compared with the \textbf{\emph{second best}} method on the large-scale popular NLPR-test, DUT-test, STRERO797 datasets, the minimum percentage gain reaches $3.0\%$, $15.3\%$, $10.7\%$ for MAE scores, respectively. 
On the small-scale LFSD dataset, compared with the \textbf{\emph{second best}} model, the percentage gain reaches $1.6\%$ in terms of F-measure and $5.0\%$ in terms of MAE score.

In order to better illustrate the advantages of our method, we analyze and summarize the qualitative and quantitative results from the following aspects:

For some common scenes, such as scenes with obvious foreground-background color contrast, large-size object, single object, simple structure, \etc., although most of the existing methods can also achieve good results, our method is more stable and robust. As shown in the first two images of Fig. \ref{visual}, the salient object is simple in structure and its color contrasts sharply with the background. In this case, while most of the works can effectively locate the salient object, our work is able to obtain more accurate results, such as sharp object boundaries (\eg, the pointy tip of the leaf in the first image), clean background suppression (\eg, the leaves in the second image).

\begin{table}[!t]
  \begin{center}
  \small
  \caption{Quantitative comparison of different methods in challenging scenes.} \label{tab_challenging}
  \renewcommand\arraystretch{1.2}
  \setlength{\tabcolsep}{1.5mm}{
  \begin{tabular}{c|c|cccc|c}
  \hline\hline
  & \multirow{2}{*}{Metrics}   & DANet  & SSF   & BiANet    & JL-DCF    & Ours  \\ 
  &   &\cite{DANet} &\cite{SSF} &\cite{BiANet} &\cite{JL-DCF} & -             \\\hline
\multirow{3}{*}{\textbf{No.1}}
  & MAE$ \downarrow$            & .0561	& .0581	& .0718	& .0636	& .0449  \\
  & $ F_{\beta}\uparrow$        & .8619	& .8608	& .8255	& .8400	& .9022  \\
  & $ S_{\alpha}\uparrow$       & .8711	& .8660	& .8455	& .8588	& .9039  \\  \hline
\multirow{3}{*}{\textbf{No.2}}
  & MAE$ \downarrow$            & .0914	& .0879	& .1503	& .1603	& .0878  \\
  & $ F_{\beta}\uparrow$        & .8495	& .8579	& .7161	& .7150	& .8715  \\
  & $ S_{\alpha}\uparrow$       & .8079	& .8139	& .7005	& .6871	& .8270  \\  \hline
\multirow{3}{*}{\textbf{No.3}}
  & MAE$ \downarrow$            & .0742	& .0603	& .0722	& .0603	& .0438  \\
  & $ F_{\beta}\uparrow$        & .8893	& .9162	& .8935	& .9114	& .9438  \\
  & $ S_{\alpha}\uparrow$       & .8691	& .8895	& .8742	& .8936	& .9225  \\  \hline
\multirow{3}{*}{\textbf{No.4}}
  & MAE$ \downarrow$            & .0400	& .0343	& .0393	& .0375	& .0279  \\
  & $ F_{\beta}\uparrow$        & .7644	& .8113	& .7742	& .7881	& .8561  \\
  & $ S_{\alpha}\uparrow$       & .8487	& .8717	& .8520	& .8597	& .9008  \\  \hline\hline
  \end{tabular}}
  \end{center}
  \end{table}

In addition, to verify the robustness and performance of our method on the challenging scenes, we conduct several sensitive studies on the testing subsets. The quantitative comparison results are shown in Table \ref{tab_challenging}.

(1) Our method has certain advantages when dealing with unreliable depth maps. As shown in Table \ref{tab_challenging} (No.1), we conduct a sensitive experiment to evaluate the performance of our method on unreliable depth map samples. Specifically, we select depth maps with the depth confidence $\lambda_d$  \cite{DCMC} score less than 0.1 from the six testing datasets as unreliable depth maps, denoted as unreliable-depth subset. 
As reported in Table \ref{tab_challenging} (No.1), compared with the \textbf{\emph{second best}} method (\ie, DANet\cite{DANet}), the percentage gain reaches 20.0\%, 4.7\%, and 3.8\% for MAE score, F-measure, and S-measure, respectively.
Moreover, as shown in the third and fourth images of Fig. \ref{visual}, the depth values of the salient objects are similar to the background, which greatly interferes with the detection of salient objects. Due to the interference of unreliable depth information, most works (\eg, S2MA\cite{S2MA}, A2dele\cite{A2dele}, D3Net\cite{D3Net}) fail to suppress the background noise, leading to the inaccurate results. Benefiting for the overall network architecture and effective cross-modality interactions,  our model can obtain robust results in the face of these unreliable factors.

(2) Our method has certain advantages when dealing with multi-object scenes. To be specific, we collect all samples with multiple salient objects from the six testing datasets based on the ground truth, denoted as multi-object subset. As shown in Table \ref{tab_challenging} (No.2), the percentage gain in both F-measure and S-measure reaches 1.6\% compared with the \textbf{\emph{second best}} method. In addition, as can be seen from the fifth and sixth images of Fig. \ref{visual}, benefiting from the cross-modality feature refinement in a global perspective, our method can not only correctly locate all salient objects, but also obtain a complete and consistent structure, such as the inner area of the person on the right in the sixth image.

(3) Our method has certain advantages when dealing with low-contrast scenes. Similarly, we select all low-contrast samples with the average color similarity between the salient objects and backgrounds exceeding 80\% from the six testing datasets (denoted as low-contrast subset) to verify the superiority of our method in this case. As shown in Table \ref{tab_challenging} (No.3), compared with the \textbf{\emph{second best}} method (\ie, SSF), the percentage gain reaches 25.9\%, 3.0\%, and 3.7\% for MAE score, F-measure, and S-measure, respectively. 
As shown in the seventh and eighth images of Fig. \ref{visual}, most of the existing works disturbed by the low color contrast interference, failing to obtain a complete result. In contrast, our method handles such a challenging scene by better exploiting complementary information across modalities, resulting in more complete and accurate results, such as hand regions of the person.

(4) Our method has certain advantages when dealing with small-object scenes. Experimentally, we select all samples with the salient object occupying less than 10\% of the image from the six testing datasets (denoted as small-object subset) to measure the performance of the proposed model in small object scenes. In Table \ref{tab_challenging} (No.4), compared with the \textbf{\emph{second best}} method (\ie, SSF),the percentage gain reaches 18.7\%, 5.5\%, and 3.3\% for MAE score, F-measure, and S-measure, respectively. As can be seen from the ninth and tenth images in Fig \ref{visual}, our method can effectively locate the small salient object, obtaining results with accurate locations, clean backgrounds, and sharp boundaries.
\begin{table}[!t]
\begin{center}
\small
\caption{Ablation studies on the NJUD-test, STEREO797 and LFSD datasets.}\label{tab2}
\renewcommand\arraystretch{1.2}
\setlength{\tabcolsep}{1.7mm}{
\begin{tabular}{c|c c|c c|c c}
\hline\hline
\multirow{2}{*}{} & \multicolumn{2}{c|}{NJUD-test} & \multicolumn{2}{c|}{STEREO797} & \multicolumn{2}{c}{LFSD}\\ \cline{2-7}
                  &$ F_{\beta}\uparrow$  & $S_{\alpha}\uparrow$   &$ F_{\beta}\uparrow$  & $S_{\alpha}\uparrow$   &$ F_{\beta}\uparrow$  & $S_{\alpha}\uparrow$   \\ \hline
Baseline          & .8880	& .8952	    & .8769	& .8865	    & .8428	& .8502  \\ \hline
+PAI              & .8952	& .8997	    & .8853	& .8937	    & .8488	& .8532  \\ \hline
+IGF              & .9135	& .9144	    & .9004	& .9064	    & .8705	& .8709  \\ \hline
+cmWR             & .9175	& .9168	    & .9062	& .9097	    & .8799	& .8713  \\ \hline
+smAR             & .9277	& .9250	    & .9139	& .9166	    & .8828	& .8753  \\ \hline\hline
\end{tabular}}
\end{center}
\end{table}

\subsection{Ablation Study}\label{4.3}

\begin{figure*}[t]
  \centering
    \includegraphics[width=1\linewidth]{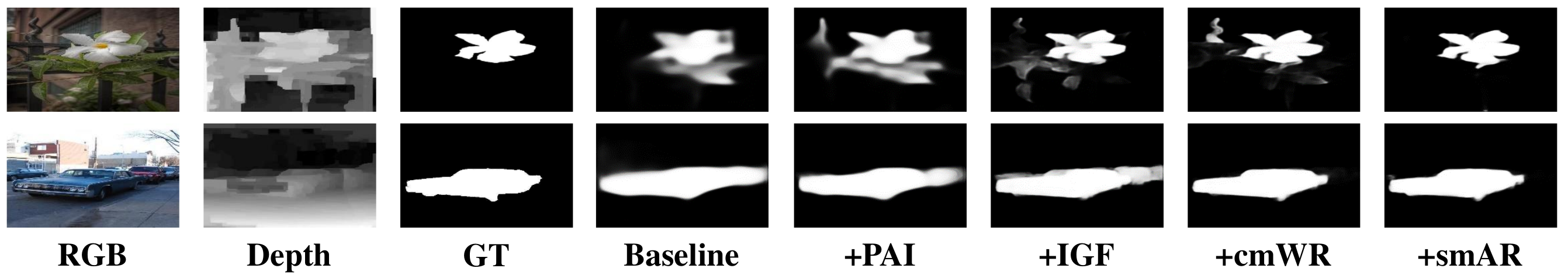}
  \caption{Visual examples of different ablation models.}
  \label{AblationVisual}
  \end{figure*}

\subsubsection{Analysis of Different Modules}
To evaluate the effectiveness of each module in the proposed model, we conduct the ablation studies on the NJUD-test, STEREO797 and LFSD datasets. The quantitative evaluations and visual examples are shown in Table \ref{tab2} and Fig. \ref{AblationVisual}, respectively.
We construct our baseline model by simplifying our full model as follows:
\begin{itemize}
	\item replacing the PAI unit with the feature concatenation of the fifth layer in the RGB and depth streams;
	\item removing the refinement middleware structure including its smAR unit and cmWR unit;
	\item replacing the IGF unit with the simple deconvolutional layers.
\end{itemize}
We use the method of progressively adding designed modules for ablation experiments. We first introduce the PAI unit into the baseline model (denoted as `+PAI'), then we progressively add the IGF unit, cmWR unit, and smAR unit into the model. In other words, `+IGF' denotes the `baseline+PAI+IGF', and the like. Moreover, all the ablation models are trained by using the same training configurations as our CIR-Net.

In Fig. \ref{AblationVisual}, it shows that the baseline model roughly locates the salient objects but lacks complete structure and sharp boundary, and many background regions are not effectively suppressed.
Compared with the baseline model, the introduction of the PAI module obtains more complete and consistent structural information (\eg, the flower in the first image), but still include many wrongly detected background regions.  From the quantitative result, the F-measure is improved from $0.8880$ to $0.8952$ on the NJUD-test dataset, and the F-measure is increased from $0.8769$ to $0.8853$ on the STEREO797 dataset.
Then, after adding the IGF unit for cross-modality feature integration in the decoder stage, the clearer boundaries of the salient objects (\eg, the flower in the first image) can be obtained and the quantitative performance is obviously improved.
Specifically, the F-measure score is increased to $0.9135$ on the NJUD-test dataset, and the percentage gain of the F-measure score reaches $2.0\%$ compared with the `+PAI' model.
Furthermore, by introducing the cmWR unit to refine the different modalities from a global perspective, it is observed that the background suppression and object structure are improved to a certain extent.
Finally, after adding the smAR unit to highlight the important cues from the single modality perspective, the full model (\ie, `+smAR' in Fig. \ref{AblationVisual} and Table \ref{tab2}) yields the best performance with the percentage gain of $4.5\%$ and $4.2\%$ in terms of F-measure on the NJUD-test dataset and STEREO797 dataset compared with the baseline model.
In summary, the ablation studies further demonstrate the effectiveness of the proposed modules.
\begin{table}[!t]
\begin{center}
\small
\caption{Quantitative comparisons of the converged three-stream architecture on the NJUD-test, STEREO797 and LFSD datasets.} \label{tab_TwoStructure}
\renewcommand\arraystretch{1.2}
\setlength{\tabcolsep}{1.7mm}{
\begin{tabular}{c|c c|c c|c c}
\hline\hline
\multirow{2}{*}{} & \multicolumn{2}{c|}{NJUD-test} & \multicolumn{2}{c|}{STEREO797} & \multicolumn{2}{c}{LFSD}\\ \cline{2-7}
		&$ F_{\beta}\uparrow$  & $S_{\alpha}\uparrow$ & $ F_{\beta}\uparrow$  & $S_{\alpha}\uparrow$ &  $ F_{\beta}\uparrow$  & $S_{\alpha}\uparrow$   \\ \hline
Full model          & .9277   & .9250  & .9139  & .9166  & .8828  & .8753  \\ \hline\hline
Two stream    & .9148   & .9134  & .9027  & .9084  & .8619  & .8568  \\ \hline\hline
RGB branch    & .8992   & .9002  & .9026  & .9089  & .8324  & .8339  \\ \hline
Depth branch  & .8450   & .8600  & .7416  & .7724  & .7681  & .7782  \\ \hline\hline

\end{tabular}}
\end{center}
\end{table}
\subsubsection{Analysis of the converged three-stream architecture}
\begin{figure}
\centering
	\includegraphics[width=1\linewidth]{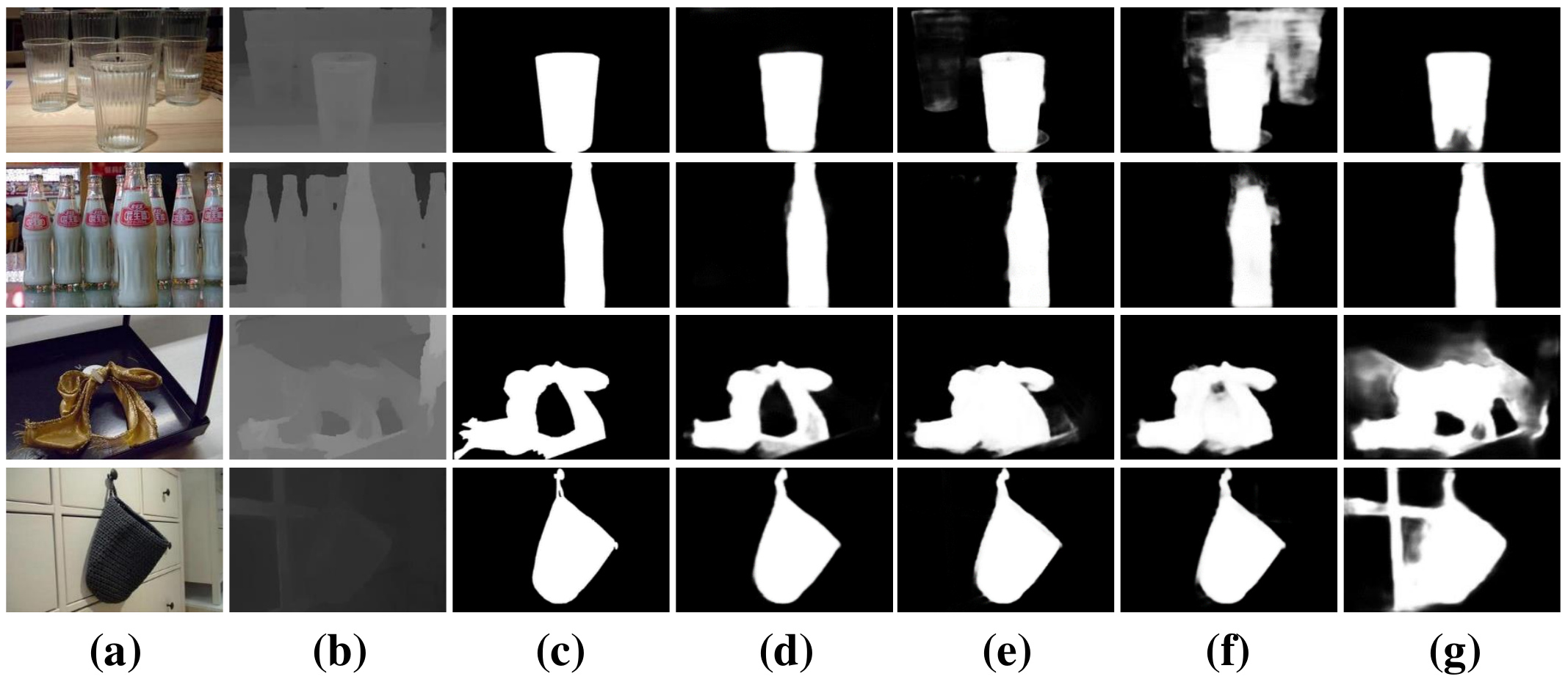}
\caption{Visual examples of two-stream structure and different branches. (a) RGB image. (b) Depth map. (c) Ground truth. (d) Our result. (e)-(g) Results of Two-stream structure, RGB branch, and Depth branch.}
\label{Two-stream}
\end{figure}

To demonstrate the effectiveness of the converged three-stream architecture, we conduct several experiments in Table \ref{tab_TwoStructure} and Fig. \ref{Two-stream}. 

First, we remove the RGB-D branch in the decoder from the full model and fuse the output features of RGB and depth branches via concatenation to obtain the final saliency map, thereby constructing the two-stream architecture network (denoted as `Two-stream'). 
From the quantitative results, we can see that, with the help of comprehensive feature interaction in the three-stream structure, the CIR-Net is more effective than the two-stream architecture. For example, on the LFSD dataset, the F-measure of the three-stream network is 0.0209 higher than that of the two-stream network, and the S-measure is 0.0185 higher. 
Similarly, from the visualization results shown in Fig. \ref{Two-stream}, we can see the advantages of the three-stream structure in detection accuracy and completeness. Of course, the performance gain comes at a price. Compared with the two-stream structure, the three-stream design needs more computational resources and parameters due to the use of more branches. To be specific, due to the additional parameters, the inference speed for an image of the three-stream architecture is 14 fps, while that of the two-stream architecture is 18 fps.

In addition, we also quantify the saliency performance of the three branches separately.
As can be seen from Fig. \ref{Two-stream}, the RGB branch and the Depth branch have their own advantages and disadvantages in different regions, but our final RGB-D branch can concentrate on the advantages of both and suppress the disadvantages, so as to achieve better results with sharper edges and complete structure.
For the quantitative comparison, it can be found that, with the help of effective cross-modality feature interaction, compared with the RGB branch performance, the final RGB-D saliency performance is significantly improved. For example, compared with the RGB branch on the LFSD dataset, the F-measure is improved from 0.8324 to 0.8828 with a percentage gain of 6.0\%, and the S-measure is improved from 0.8339 to 0.8753 with a percentage gain of 5.0\%. 
These experiments demonstrate the robustness and effectiveness of the proposed model architecture.

\begin{table}[!t]
\begin{center}
\small
\caption{Ablation studies on middleware on the NJUD-test, STEREO797 and LFSD datasets.} \label{tab_MiddleWare}
\renewcommand\arraystretch{1.4}
\setlength{\tabcolsep}{1.2mm}{ 
\begin{tabular}{c|c|c c|c c|c c}
\hline\hline
\multicolumn{2}{c|}{\multirow{2}{*}{}} & \multicolumn{2}{c|}{NJUD-test} & \multicolumn{2}{c|}{STEREO797} & \multicolumn{2}{c}{LFSD}	\\ \cline{3-8}
\multicolumn{2}{c|}{} &$ F_{\beta}\uparrow$  & $S_{\alpha}\uparrow$ & $ F_{\beta}\uparrow$  & $S_{\alpha}\uparrow$ &  $ F_{\beta}\uparrow$  & $S_{\alpha}\uparrow$   \\ \hline
\multicolumn{2}{c|}{Full model} & .9277   & .9250  & .9139  & .9166  & .8828  & .8753  \\ \hline\hline
\multirow{3}{*}{\rotatebox{90}{smAR}}  	
& w/CA, w/o SA 	  & .9152   & .9143  & .9041  & .9090  & .8782  & .8725  \\ \cline{2-8}
& w/o CA, w/SA 	  & .9114   & .9119  & .9031  & .9085  & .8785  & .8689  \\ \cline{2-8}
& SA-CA	  	  & .9167   & .9146  & .8997  & .9030  & .8611  & .8549  \\ \hline\hline
\multirow{2}{*}{\rotatebox{90}{cmWR}} 
& w/ M1, w/o M2      & .9135   & .9114  & .9008  & .9004  & .8623  & .8595  \\ \cline{2-8}
& w/o M1, w/ M2  	  & .9090   & .9085  & .8959  & .9004  & .8582  & .8462  \\ \hline\hline
\end{tabular}}
\end{center}
\end{table}

\subsubsection{Analysis of Refinement Middleware}
We conduct various ablation experiments in Table \ref{tab_MiddleWare} to validate the effectiveness of the refinement middleware. 

In terms of the proposed smAR unit, we replace the 3D attention tensor with a single channel attention weight (denoted as ‘w/ CA, w/o SA’), a single spatial attention weight (denoted as ‘w/o CA, w/ SA’), and a serial utilization of the SA-CA combination (denoted as ‘SA-CA’). 
From Table \ref{tab_MiddleWare}, we can see that the proposed smAR unit is more effective than other commonly used attention variants. 
For example, compared with the serial utilization of SA-CA combination module on the STEREO797 dataset, the F-measure of our full model with smAR unit reaches 0.9139 with a percentage gain of 1.6\%, and the percentage gain of S-measure is 1.5\%.

In terms of the proposed cmWR unit, we replace the final weight map (\ie, $M_1 \times M_2$) with the $M_1$-only (denoted as `w/ M1, w/o M2') and $M_2$-only (denoted as `w/o M1, w/ M2') cases to demonstrate the advantages of the cmWR unit. 
From the quantitative comparison reported in Table \ref{tab_MiddleWare}, we can see that the way of $M_1 \times M_2$ is more effective than the single weight map $M_1$ or $M_2$. 
For example, on the LFSD dataset, compared with the case of only $M_2$, the F-measure score is improved from 0.8582 to 0.8828 with a percentage gain of 2.9\% and the S-measure score is improved from 0.8462 to 0.8753 with a percentage gain of 3.4\%.

\subsubsection{Analysis of different feature interaction strategy in PAI and IGF units}
To verify the effectiveness of our design of PAI and IGF units, we conduct various experiments, as shown in Table \ref{tab_PAI+IGF}. 

In terms of the PAI unit, we add two ablation experiments. One is to validate the combination and propagation of different layers, and the other is to replace the spatial attention maps with the fused cross-modality RGB-D features in the corresponding layer (denoted as `Trans Fusion’). 
From Table \ref{tab_PAI+IGF}, we can see that starting combination from the third layer (the Full model) achieves the best performance, and the PAI unit is more effective than the commonly used feature fusion strategy.
For example, on the LFSD dataset, the F-measure score is improved from 0.8555 to 0.8828 with a percentage gain of 3.2\% compared with the forward propagation from the first layer, and the S-measure score is improved from 0.8480 to 0.8753 with a percentage gain of 3.2\% compared with the feature fusion strategy.

\begin{table}[!t]
\begin{center}
\small
\caption{Ablation studies on PAI and IGF units on the NJUD-test, STEREO797 and LFSD datasets.} \label{tab_PAI+IGF}
\renewcommand\arraystretch{1.2}
\setlength{\tabcolsep}{1.2mm}{ 
\begin{tabular}{c|c|c c|c c|c c}
\hline\hline
\multicolumn{2}{c|}{\multirow{2}{*}{}} & \multicolumn{2}{c|}{NJUD-test} & \multicolumn{2}{c|}{STEREO797} & \multicolumn{2}{c}{LFSD}	\\ \cline{3-8}
\multicolumn{2}{c|}{} &$ F_{\beta}\uparrow$  & $S_{\alpha}\uparrow$ & $ F_{\beta}\uparrow$  & $S_{\alpha}\uparrow$ &  $ F_{\beta}\uparrow$  & $S_{\alpha}\uparrow$   \\ \hline
\multicolumn{2}{c|}{Full model} & .9277   & .9250  & .9139  & .9166  & .8828  & .8753  \\ \hline\hline
\multirow{5}{*}{\rotatebox{90}{PAI}}   	
& Fifth Layer 	  & .9098	& .9120	 & .9033	& .9063   & .8318	& .8345     \\ \cline{2-8}
& Fourth Layer 	  & .9081	& .9130	 & .8962	& .8981	  & .8657	& .8546     \\ \cline{2-8}
& Second Layer	  & .9105	& .9134	 & .9036	& .9069	  & .8620	& .8544     \\ \cline{2-8}
& First Layer	 	  & .9104	& .9123	 & .9036	& .9036	  & .8555	& .8487  	\\ \cline{2-8}\cline{2-8}
& Trans Fusion	  & .9123	& .9143	 & .9034	& .9094	  & .8534	& .8480     \\ \hline\hline
\multirow{2}{*}{\rotatebox{90}{IGF}}
& w/ add      & .9106	& .9131	 & .9052	& .9086	  & .8609	& .8530     \\ \cline{2-8}
& w/ cat  	  & .9132	& .9151	 & .9039	& .9082	  & .8605	& .8518     \\ \hline\hline
\end{tabular}}
\end{center}
\end{table}	

\begin{table}[!t]
\begin{center}
\small
\caption{Ablation study of the addition operation in Eq. (\ref{eq2}, \ref{eq5}, \ref{eq8}) on NJUD-test, STEREO797, and LFSD datasets.} \label{tab_w/o residual connection}
\renewcommand\arraystretch{1.2}
\setlength{\tabcolsep}{1.2mm}{
\begin{tabular}{c|c|c|c|c|c|c}
\hline\hline
\multirow{2}{*}{} & \multicolumn{2}{c|}{NJUD-test} & \multicolumn{2}{c|}{STEREO797} & \multicolumn{2}{c}{LFSD}\\ \cline{2-7}
		&$ F_{\beta}\uparrow$  & $S_{\alpha}\uparrow$ & $ F_{\beta}\uparrow$  & $S_{\alpha}\uparrow$ &  $ F_{\beta}\uparrow$  & $S_{\alpha}\uparrow$   \\ \hline
Full model          			& .9277   & .9250  	& .9139  & .9166  	& .8828  & .8753     \\ \hline\hline
w/o add in Eq. (\ref{eq2})	& .9171	  & .9159	& .9058	 & .9082	& .8500	 & .8468	 \\ \hline
w/o add in Eq. (\ref{eq5})	& .9100	  & .9114	& .9023	 & .9047	& .8606	 & .8490	 \\ \hline
w/o add in Eq. (\ref{eq8})	& .9096	  & .9113	& .9064	 & .9071	& .8631	 & .8552     \\ \hline\hline
\end{tabular}}
\end{center}
\end{table}

In terms of IGF unit, we replace the dynamic fusion strategy with the addition (denoted as ‘w/ add’) or concatenation (denoted as ‘w/ cat’) to suggest the effectiveness of the IGF unit. 
From Table \ref{tab_PAI+IGF}, we can see that with help of the proposed IGF unit, the performance is improved when compared with the commonly used fusions strategy (add or concatenation). 
For example, on the LFSD dataset, compared with the concatenation operation (\ie, w/ cat), the F-measure score is improved from 0.8605 to 0.8828 with a percentage gain of 2.6\% and the S-measure score is improved from 0.8518 to 0.8753 with a percentage gain of 2.8\%. 

\subsubsection{Analysis of residual connection}
To demonstrate the effectiveness of residual features, we conduct the ablation studies that remove the addition operation in Eqs. (\ref{eq2}, \ref{eq5}, \ref{eq8}). The quantitative results are shown in Table \ref{tab_w/o residual connection}. 
Compared with the only direct multiplication operation, our method with residual connection achieves better quantitative performance. 
For example, in Table \ref{tab_w/o residual connection}, compared with the result of removing the addition in Eq. (\ref{eq2}), on the LFSD dataset, the F-measure is improved from 0.8500 to 0.8828 with a percentage gain of 3.9\% and the percentage gain of S-measure reaches 3.4\%. 
Similarly, removing the addition operations in Eq. (\ref{eq5}, \ref{eq8}) also degrades the performance.

\subsubsection{Analysis of effectiveness to different scenes}
When the depth map is unreliable, through the cross-modality interaction of PAI unit in the encoder stage, the features of RGB-D branch can exploit the correlation between RGB and depth modalities to highlight the salient regions. Moreover, in the decoder stage, the IGF unit is able to selectively determine the contribution of depth modality, thus suppressing the interference of unreliable information in depth modality. 
To validate the effectiveness of the PAI and IGF units on unreliable depth maps, we add two ablation studies on the unreliable-depth subset that replace the PAI unit with the feature concatenation of the fifth layer in the RGB and depth branches (denoted as `w/o PAI’), and replace the IGF unit with the direct feature concatenation (denoted as `w/o IGF’), respectively. As shown in the left side of Table \ref{tab_ablation_challenging}, it can be found that after removing the PAI unit and the IGF unit, the detection effect of the model on unreliable depth maps decreases. For example, with the PAI unit, the F-measure is improved from 0.8886 to 0.9022 with a percentage gain of 1.5\%. Similarly, with the IGF unit, the F-measure is improved from 0.8978 to 0.9022 with a percentage gain of 0.5\%.

\begin{table}[!t]
\begin{center}
\scriptsize
\caption{Ablation study of different modules in unreliable depth maps and multiple-object scenes.} \label{tab_ablation_challenging}
\renewcommand\arraystretch{1.5}
\setlength{\tabcolsep}{1.2mm}{
\resizebox{8cm}{!}{
\begin{tabular}{c|c|cc||c|c}
\hline\hline
 &\multicolumn{3}{c||}{unreliable-depth subset} & \multicolumn{2}{c}{multi-object subset} \\ \hline
Metrics   &Ours  &w/o PAI    &w/o IGF    &Ours  &w/o cmWR  \\ \hline
$ F_{\beta}\uparrow$    &.9022	&.8886	&.8978  &.8715	&.8626	\\
$ S_{\alpha}\uparrow$   &.9039	&.8932	&.8992  &.8270	&.8255	\\ \hline\hline
\end{tabular}}}
\end{center}
\end{table}

In addition, concerning the challenging scene containing multiple salient objects, the cmWR unit can extract the cross-modality global context information by calculating the long-range dependency, thus refining features from a global perspective and improving the completeness of saliency results. Similarly, we add an ablation experiment to demonstrate the effectiveness of the cmWR unit in the multi-object scene. As can be reported in the right side of Table \ref{tab_ablation_challenging}, compared with the model without cmWR unit (denoted as ‘w/o cmWR’), the F-measure is improved from 0.8626 to 0.8715 with a percentage gain of 1.0\%.

\subsection{Discussion}\label{4.4}     
\subsubsection{Failure Cases}\label{Failure Cases Section}
\begin{figure}[t]
\centering
	\includegraphics[width=1\linewidth]{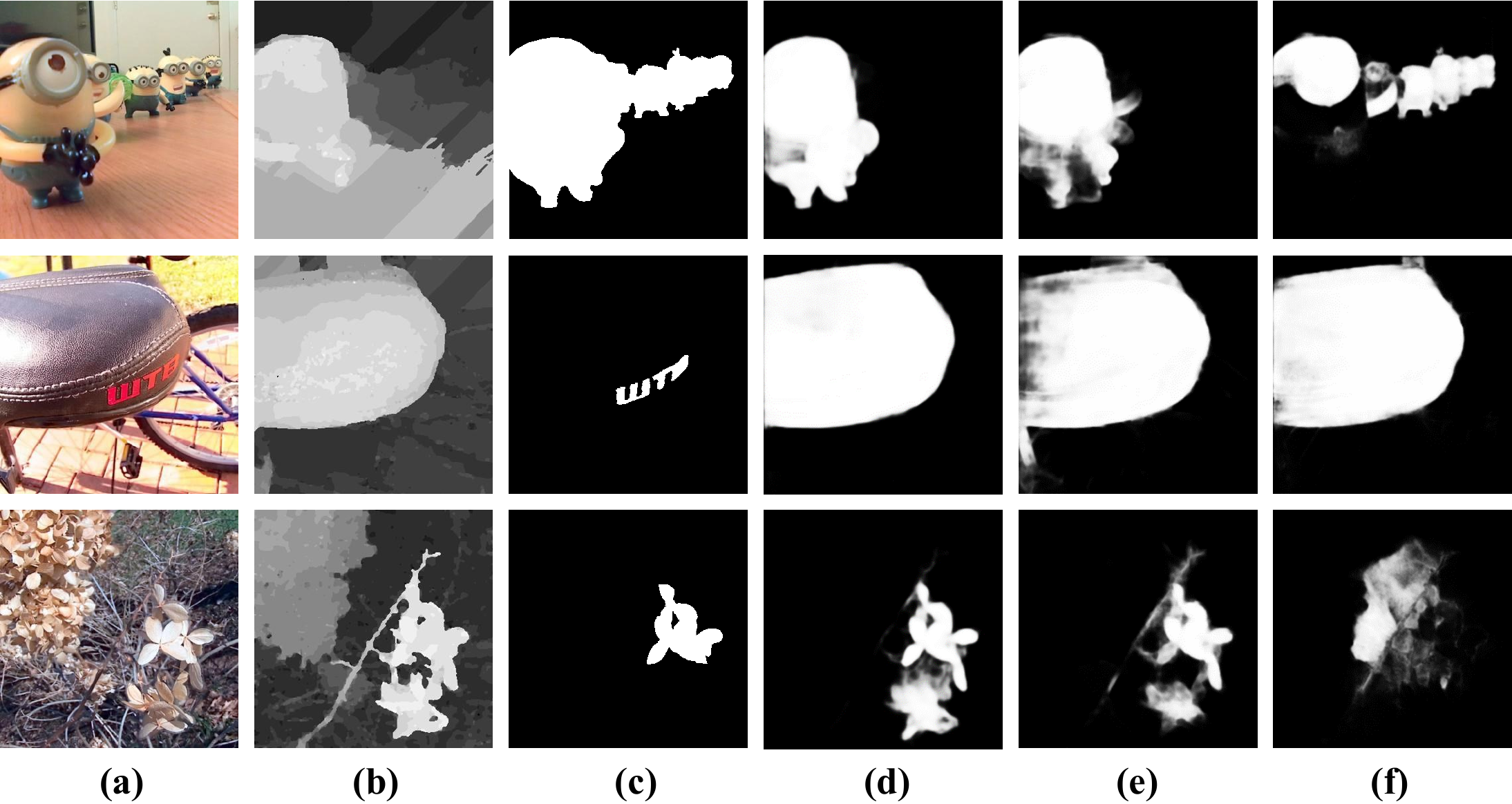}
\caption{Visual examples of failure cases. (a) RGB image. (b) Depth map. (c) GT. (d) Ours. (e) S2MA \cite{S2MA}. (f) DANet \cite{DANet}.}
\label{failure_cases}
\end{figure}

Several representative failure cases are shown in Fig. \ref{failure_cases}. We can see that it is difficult to perfectly locate salient objects in the following aspects: 
1) Multiple and small salient objects. 
In the first scene, although the multiple salient objects contain the same characteristics in the scene, the salient objects far from the lens are too small, so that the corresponding depth map fails to provide effective depth information of these objects. 
Hence, it is difficult to completely detect all salient objects in such a scene. 
2) High contrast but not salient objects. 
In the second scene, it is obvious that the bike seat contrasts sharply with the background in the depth map. 
However, the red logo, the real salient object, is also in sharp contrast to the bike seat in the RGB image. 
Therefore, the ambiguity introduced by this conflict prevents our model from accurately detecting the red logo as the salient object.
3) Complex background noise. 
In the third scene, due to the small contrast between the salient object and the background of the RGB image and the misleading depth information in the depth map, our algorithm fails to suppress the background effectively. 
It is worthy to note that, for the above challenging scenes, the recent state-of-the-art methods (S2MA \cite{S2MA} and DANet \cite{DANet}) also fail to detect the salient objects correctly. 

\subsubsection{Future Work}
In the future, work in three areas can be further studied. 
First, our paper mainly focuses on how to achieve cross-modality interaction more fully and effectively and does not specifically consider the solution when the quality of the depth map is unreliable, but only uses some control mechanisms (\eg, cmWR and IGF modules) to reduce the negative impact of low-quality depth maps. 
Under the existing depth imaging equipment, how to stably and explicitly achieve salient object detection in the case of poor depth map quality is a problem worthy of study.
Second, as we all know, deep learning-based methods are data-driven. Thus, more training data would improve the generalization capability of deep models in most cases.
As presented in Table \ref{tab_w/o NLPR}, when discarding the training data of the NLPR dataset (\ie, `w/o NLPR'), the final performance is all degraded, but to varying degrees. 
For example, on the STEREO797 dataset, the F-measure drops by only 0.5\%, but on the LFSD dataset it drops by 4.4\%. 
Put like that, constructing larger datasets or reducing the dependence on data volume under the premise of ensuring performance can be worked as future research directions. 
Furthermore, weakly supervised RGB SOD task has received a lot of attention, but very little in RGB-D SOD. 
Exploring RGB-D SOD models with less supervisory information can reduce the dependence on data annotation and is a very valuable and promising research direction. 
Last but not least, two- and three-stream based RGB-D SOD models have achieved satisfactory performance, but as a fundamental pre-processing task, how to pursue real-time efficiency while maintaining performance is also a valuable research point.

\begin{table}[!t]
{
\begin{center}
\small
\caption{Quantitative evaluation of our method without NLPR-train dataset on NJUD-test, STEREO797, and LFSD datasets.} \label{tab_w/o NLPR}
\renewcommand\arraystretch{1.2}
\setlength{\tabcolsep}{1.5mm}{
\begin{tabular}{c|c|c|c|c|c|c}
\hline\hline
\multirow{2}{*}{} & \multicolumn{2}{c|}{NJUD-test} & \multicolumn{2}{c|}{STEREO797} & \multicolumn{2}{c}{LFSD}\\ \cline{2-7}
		&$ F_{\beta}\uparrow$  & $S_{\alpha}\uparrow$ & $ F_{\beta}\uparrow$  & $S_{\alpha}\uparrow$ &  $ F_{\beta}\uparrow$  & $S_{\alpha}\uparrow$   \\ \hline
Full model          & .9277   & .9250  & .9139  & .9166  & .8828  & .8753  \\ \hline\hline
w/o NLPR	  & .9170 	& .9177	 & .9096  & .9129  & .8436	& .8488	 \\ \hline\hline
\end{tabular}}
\end{center}}
\end{table}
\section{Conclusion}\label{sec5}

In this work, we proposed an end-to-end network, named CIR-Net, for the task of RGB-D SOD.
The strength of our algorithm comes from the synergy of the model architecture and technical modules.

From the perspective of model architecture, we design a new three-stream-like model architecture to more comprehensively realize cross-modality information interaction. As we all know, the two-stream models are currently the most widely used structure in RGB-D SOD task, mainly including an RGB branch and a depth branch, which can achieve the cross-modality interaction in the feature encoder or decoder stage. However, the two-stream models can only complete the interaction of RGB and depth modalities, while ignoring the role of RGB-D modality. In contrast, the three-stream structure has the opportunity to model the correlation and interaction among the RGB, depth, and RGB-D modalities. Moreover, our proposed model architecture is also different from the existing three-stream structures. On the one hand, the generation of our RGB-D stream is not learned from scratch, but obtained through the fusion of the high-level features from the RGB branch and depth branch through the PAI module, which can make the learned RGB-D features more discriminative and reduce the amount of calculation. On the other hand, we adopt a clear convergence structure at the decoder stage to realize the information interaction centered on the RGB-D modality, which can further capture the complementarity of the three modalities (\ie, RGB, depth, and RGB-D), thereby obtaining more discriminative and saliency-related features. 

From the technical design level, as our title says, we do two things in this paper: cross-modality interaction and cross-modality refinement. For the cross-modality interaction, different from the existing cross-modality interaction methods that operated only in the encoder or decoder stage, we dedicate to integrating cross-modality information into both encoder and decoder stages jointly in a more comprehensive and in-depth manner. Concretely, in the feature encoder stage, a PAI unit is designed to fuse the cross-modality and cross-level features, thereby attaining the RGB-D encoder representations. In the feature decoder stage, we design a convergence structure equipped with the IGF unit to make the RGB and depth decoder features flow into the RGB-D mainstream branch, and effectively select the most valuable supplementary information from RGB and depth modalities to obtain more discriminative cross-modality saliency prediction features. For the cross-modality refinement, we insert a refinement middleware between the encoder and decoder to further highlight the effective information before decoding from the perspective of self-modality and cross-modality. Specifically, we propose a simple but effective smAR unit in a 3D-tensor manner to reduce the feature redundancy of the channel dimension and emphasize the important location of the spatial dimension, as well as propose a cmWR unit to refine the multi-modality features by considering cross-modality complementary information and cross-modality global contextual dependencies. It is worth mentioning that such a middleware structure is pluggable for three-stream networks.

The mutual cooperation and facilitation of model structure and technical modules enable our method to achieve competitive performance on six datasets both qualitatively and quantitatively.

\par
\ifCLASSOPTIONcaptionsoff
  \newpage
\fi
{
\bibliographystyle{IEEEtran}
\bibliography{ref}
}

\end{document}